\documentclass[10pt,twocolumn,letterpaper]{article}
\usepackage{multirow}
\usepackage[misc]{ifsym}

\usepackage[pagenumbers]{cvpr} 

\usepackage[dvipsnames]{xcolor}

\definecolor{cvprblue}{rgb}{0.21,0.49,0.74}
\usepackage[pagebackref,breaklinks,colorlinks,citecolor=cvprblue]{hyperref}
\usepackage{xcolor}
\usepackage{pifont}
\usepackage{cases}
\usepackage{algorithm}
\usepackage{algpseudocode}
\usepackage{etoolbox}
\definecolor{best}{RGB}{0,191,255}
\definecolor{second}{RGB}{137,207,240}

\newcommand{\best}[1]{\textbf{#1}}
\newcommand{\second}[1]{\underline{#1}}

\usepackage{tcolorbox}

\algnewcommand{\LineComment}[1]{\State \(\triangleright\) #1}

\newcommand{\balpha}{\boldsymbol{\alpha}}
\newcommand{\bbeta}{\boldsymbol{\beta}}

\usepackage{amsmath,amsfonts,bm}

\def\eqref#1{equation~\ref{#1}}

\def\1{\bm{1}}

\DeclareMathAlphabet{\mathsfit}{\encodingdefault}{\sfdefault}{m}{sl}
\SetMathAlphabet{\mathsfit}{bold}{\encodingdefault}{\sfdefault}{bx}{n}

\def\gC{{\mathcal{C}}}
\def\gD{{\mathcal{D}}}
\def\gE{{\mathcal{E}}}

\def\gP{{\mathcal{P}}}

\def\gX{{\mathcal{X}}}
\def\gY{{\mathcal{Y}}}

\newcommand{\vect}[1]{\ensuremath{\mathbf{#1}}}

\makeatletter\renewcommand\paragraph{\@startsection{paragraph}{4}{\z@}
{.4em \@plus1ex \@minus.2ex}{-.5em}{\normalfont\normalsize\bfseries}}\makeatother

\BeforeBeginEnvironment{equation}{\vspace{-5pt}}
\AfterEndEnvironment{equation}{\vspace{0pt}}

\def\paperID{} 
\def\confName{CVPR}
\def\confYear{2024}

\title{Instruct2Attack: Language-Guided Semantic Adversarial Attacks}

\author{Jiang Liu\textsuperscript{1}, Chen Wei\textsuperscript{1}, Yuxiang Guo\textsuperscript{1}, Heng Yu\textsuperscript{2}, Alan Yuille\textsuperscript{1}, Soheil Feizi\textsuperscript{3}, \\ Chun Pong Lau\textsuperscript{4 \Letter}, Rama Chellappa\textsuperscript{1}\\
\textsuperscript{1}Johns Hopkins University, \textsuperscript{2}Carnegie Mellon University, \\ 
\textsuperscript{3}University of Maryland, College Park, \textsuperscript{4}City University of Hong Kong}

\begin{document}
\twocolumn[{%
\renewcommand\twocolumn[1][]{#1}%
\maketitle
\begin{center}
    \centering
    \captionsetup{type=figure}
    \includegraphics[width=0.98\textwidth]{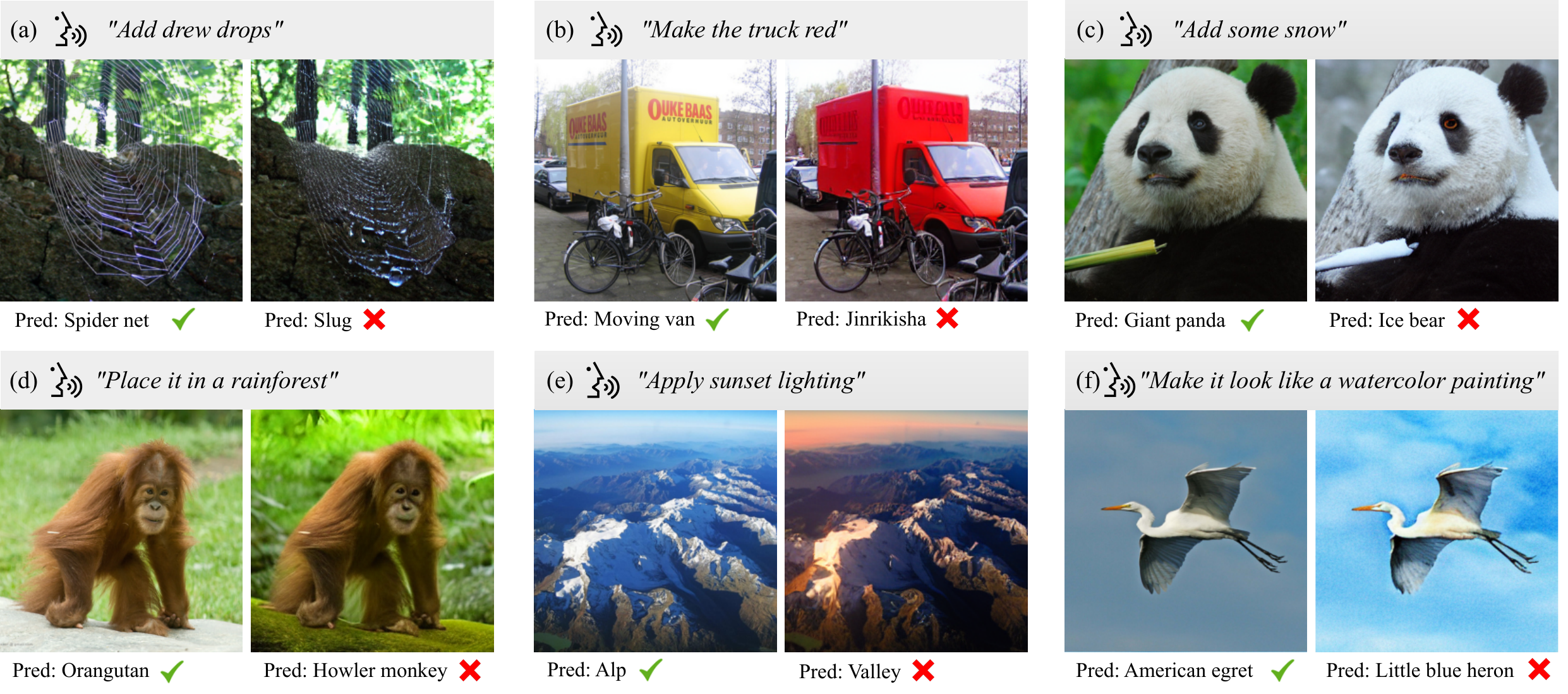}
    \captionof{figure}{
    \textbf{Examples of Instruct2Attack}. Given an input image (\textit{left}) and an edit instruction (\textit{top banner}), our method performs adversarial semantic editing (\textit{right}) on the original image according to the instruction. The edited images can successfully fool the deep classifier and effectively reveal its vulnerability with subtle semantic changes on, \eg, object shape (\textit{a}), object color (\textit{b}), weather (\textit{c}), background environment (\textit{d}), lighting condition (\textit{e}), and image style (\textit{f}).  The edit instructions here are automatically generated by GPT-4. 
    }
    \label{fig:teaser}
\end{center}%
}]

\begin{abstract}
We propose Instruct2Attack (I2A), a language-guided semantic attack that generates semantically meaningful perturbations according to free-form language instructions. We make use of state-of-the-art latent diffusion models, where we adversarially guide the reverse diffusion process to search for an adversarial latent code conditioned on the input image and text instruction. Compared to existing noise-based and semantic attacks, I2A generates more natural and diverse adversarial examples while providing better controllability and interpretability. We further automate the attack process with GPT-4 to generate diverse image-specific text instructions. We show that I2A can successfully break state-of-the-art deep neural networks even under strong adversarial defenses, and demonstrate great transferability among a variety of network architectures. 
\end{abstract} 
\section{Introduction}
Shortly after Deep Neural Networks (DNNs) began to outperform other methods in the field of computer vision~\cite{krizhevsky2012imagenet, Girshick2013RichFH, vgg}, their vulnerability with adversarial examples was also discovered~\cite{szegedy2013intriguing}, which raises crucial security concerns~\cite{akhtar2018threat}. Specifically, it is demonstrated that DNNs could be fooled by adding small but carefully crafted adversarial noises to the input images, which are imperceptible to human eyes but can cause DNNs to misclassify the images with high confidence. 

While earlier work is focused on noise-based attacks~\cite{goodfellow2014explaining,madry2018towards, carlini2017towards}, recent research explores \textit{semantically meaningful} adversarial examples~\cite{joshi2019semantic,qiu2020semanticadv}. Unlike noise-based techniques that produce subtle pixel changes, semantic attacks aim to expose systemic vulnerabilities through semantic manipulations. For instance, an image classifier could be fooled by adversarial semantic image editing that is not supposed to change the object category, such as spatial~\cite{xiao2018spatially, engstrom2018rotation, dong2022viewfool} and color transforms~\cite{, laidlaw2019functional, hosseini2018semantic, zhao2021large}, global style transformation~\cite{kang2019testing, shamshad2023clip2protect}, attribute editing~\cite{joshi2019semantic,qiu2020semanticadv} and other semantic perturbations~\cite{xu2017can,liu2018beyond}. Such adversarial examples provide additional insights into model failure modes beyond pixel-wise perturbations and address criticisms that noise-based attacks lack interpretability and naturalness, and fail to capture model failures in the real world.

A key challenge in crafting semantic adversarial examples is making controlled edits that only change a specific attribute while preserving other semantics. Previous approaches~\cite{qiu2020semanticadv, shamshad2023clip2protect} have applied generative adversarial networks (GANs) for controllable image editing~\cite{goodfellow2014generative,radford2015unsupervised,zhu2017unpaired}. However, GANs have limited disentanglement capabilities for complex real-world images beyond structured face and street-view images. It is also not easy for GAN-based image editing to arbitrarily manipulate any semantics, such as through natural language instructions. Recent advances in controllable image generation and editing with diffusion models~\cite{sohl2015deep, ho2020denoising, nichol2021improved} have opened new opportunities for designing semantic adversarial attacks. For instance, InstructPix2Pix~\cite{brooks2023instructpix2pix} enables text-driven image manipulation by conditioning on an input image and a written instruction to produce an edited result. This approach demonstrates impressive capabilities in following free-form language instructions to make targeted changes to diverse images.

Building on recent advances in controllable image editing, we propose Instruct2Attack (I2A), a novel attack method that generates realistic and interpretable semantic adversarial examples guided by free-form language instructions.
Specifically, our approach leverages a latent conditional diffusion model~\cite{rombach2022high, brooks2023instructpix2pix} which has two conditions: One is the input image to perturb, and the other is the language instruction defining semantic attributes to modify. Controlled by two learnable adversarial guidance factors corresponding to the image and text guidance respectively, I2A navigates in the latent space to find an adversarial latent code that generates semantic adversarial images for the targeted victim models. The generated adversarial examples selectively edit attributes based on the language instruction while preserving unrelated image context. Additionally, we impose a perceptual constraint to ensure the similarity between the input image and the adversarial image.

Our method reveals the vulnerability of the victim models in various situations with a simple control knob of natural languages. We further explore automating the process with GPT-4~\cite{openai2023gpt} to generate diverse and image-specific text instruction as illustrated in \cref{fig:teaser}. The proposed I2A attack introduces natural and diverse semantic perturbations based on text instructions. Quantitative experiments demonstrate that I2A delivers high attack success rates, especially under strong adversarial training-based~\cite{liu2023comprehensive, engstrom2019learning, salman2020adversarially, Wong2020Fast, singh2023revisiting} and preprocessing-based~\cite{nie2022DiffPure} defenses, where I2A consistently outperforms previous adversarial attack methods by a large margin. For example, I2A can effectively break the strongest Swin-L~\cite{liu2021swin, liu2023comprehensive} model on the Robustbench~\cite{croce2021robustbench} ImageNet leaderboard, bringing down the robust accuracy from 59.56\% to 4.98\%. Moreover, we demonstrate that I2A achieves great attack transferability among a variety of model architectures under the black-box setting.





\label{sec:intro}

\section{Related Work}
\label{sec:related}
\paragraph{$L_p$ and non-$L_p$ attack.} $L_p$ attack~\cite{goodfellow2014explaining, madry2018towards, carlini2017towards, lau2023adversarial, liu2022segment, liu2022mutual} optimizes adversarial noises within a small $L_p$ ball, which is the most popular form of adversarial attacks. $L_p$ adversarial samples are often imperceptible but they are shown to have a distribution gap from the natural images~\cite{yoon2021adversarial}. On the other hand, non-$L_p$ attack generalizes $L_p$ attack by using non-$L_p$ metrics to bound the perturbations~\cite{zhao2021large, laidlaw2021perceptual, kang2019testing} or even generating unbounded perturbations~\cite{Bhattad2020Unrestricted}, which allows large and perceptible perturbations.


\looseness -1 \paragraph{Semantic attack.} Semantic attack is a type of non-$L_p$ attack that aims to alter the semantic content or context of the input. Semantic attack creates interpretable perturbations that are more relevant in real-world scenarios. Early works in semantic attacks are limited to simple transformations including spatial transform~\cite{xiao2018spatially, engstrom2018rotation, dong2022viewfool}, color transforms~\cite{, laidlaw2019functional, hosseini2018semantic, zhao2021large}, and brightness adjustment~\cite{xu2017can}. More recent work includes global style transformation~\cite{kang2019testing, shamshad2023clip2protect}, and fixed-category attribute editing~\cite{joshi2019semantic,qiu2020semanticadv}. Compared to existing works, I2A is capable of generating more sophisticated and flexible semantic manipulations based on free-form language instructions. 
\paragraph{Latent space attack.} Instead of adding perturbations in the image space, latent space attacks perturb the latent representation of the input image, which is usually in the latent space of generative models, such as GANs \cite{jalal2017robust, lin2020dual, lau2023attribute}, flow-based models \cite{yuksel2021semantic, lau2023interpolated} and diffusion models \cite{liu2023diffprotect, xue2023diffusion, chen2023diffusion, wang2023semantic}. Latent space attacks generate more structured perturbations than image space attacks. However, these perturbations usually lack semantic meaning and are hard to interpret.
\paragraph{Image editing using generative models.}
Image editing \cite{luo2022context, ntavelis2020sesame, luo2023siedob, ling2021editgan} denotes the process of manipulating a given image by certain semantic attributes, such as style transfer and image domain translation. It requires the techniques from semantic image synthesis \cite{isola2017image, tang2020local, park2019semantic} and image inpainting \cite{yu2018generative, yu2019free, zhao2021large}. Various editing approaches first encode \cite{chai2021using, richardson2021encoding} or invert \cite{abdal2019image2stylegan, abdal2020image2stylegan++, alaluf2022hyperstyle} images to obtain the corresponding latent representation and edit the images by manipulating the latent vectors. Recently, text encoders such as CLIP \cite{radford2021learning} or BERT \cite{kenton2019bert} are also employed to guide the image editing process with text using GAN models  \cite{crowson2022vqgan, crowson2022vqgan, gal2022stylegan} or diffusion models \cite{avrahami2022blended, kim2022diffusionclip, nichol2022glide, hertz2023prompttoprompt, brooks2023instructpix2pix}.  


\begin{figure*}
    \centering
    \vspace{-7pt}
    \includegraphics[width=1.02\textwidth]{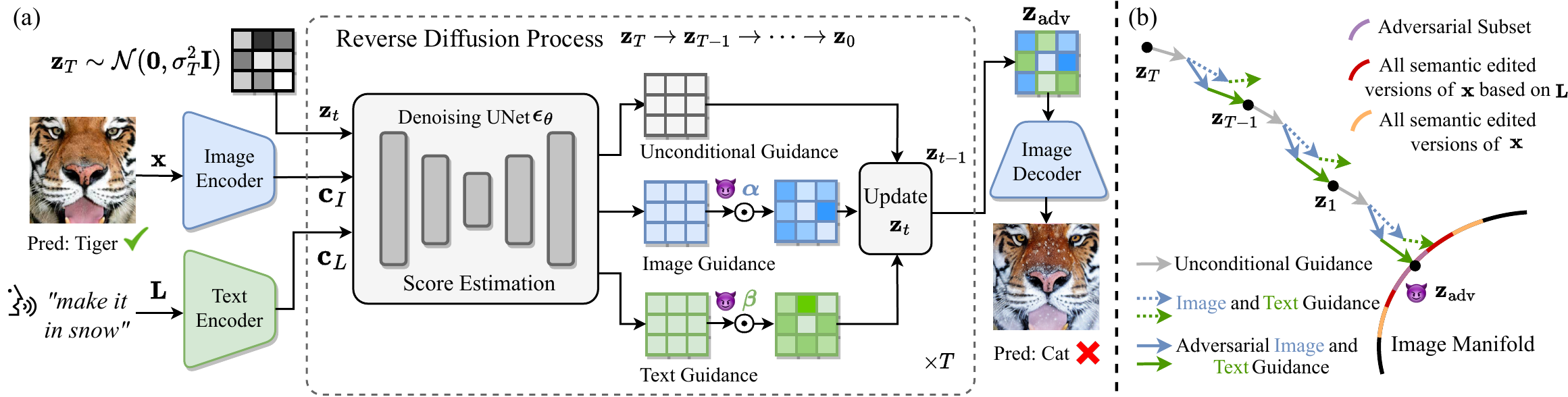}
    \vspace{-15pt}
    \caption{(a) \textbf{Overview of the Instruct2Attack (I2A) framework.} (b) \textbf{Illustration of adversarial diffusion guidance.} Given the input image and text instruction as conditions, I2A adversarially guides the reverse diffusion process to search for an adversarial code in the latent space by modulating the image and text guidance with the adversarial guidance factors $\balpha$ and $\bbeta$.}
    \label{fig:method}
    \vspace{-10pt}
\end{figure*}
\section{Method}
\vspace{-3pt}
\subsection{Problem Definition}
\vspace{-3pt}
In this paper, we consider the task of image classification. Suppose an image sample  $\vect{x}\in \gX := \mathbb{R}^{H \times W \times C}$ and its corresponding label $y\in \gY :=\{1,\cdots, |\gY|\}$ are drawn from an underlying distribution $\mathbb{P}:=\gX \times \gY$, where $H$, $W$ and $C$ are the height, width and the number of channels of the image, respectively. Let $f: \gX \rightarrow \gY$ be an image classifier. Given a clean input image $\vect{x}$ and the corresponding ground-truth label ${y}$, the goal of the attacker is to craft an adversarial image $\vect{x}_\text{adv}$ of $\vect{x}$ to fool $f$, such that $f(\vect{x}_{\text{adv}})\neq {y}$.

A popular way to construct $\vect{x}_\text{adv}$ is by adding pixel-wise adversarial noise $\delta$~\cite{goodfellow2014explaining, madry2018towards, carlini2017towards} to the input image $\vect{x}$, \ie, $\vect{x}_\text{adv}=\vect{x}+\delta$. Although it has been shown that noise-based attacks are successful in breaking DNNs, $\delta$ lacks semantic meanings and is unlikely to exist in natural images, which makes it hard to interpret why they cause the failures of DNNs. In this work, we instead consider language-guided semantic attacks, where we aim to generate semantically meaningful perturbations guided by free-form languages. These natural and interpretable perturbations provide valuable insights into understanding the failure modes of DNNs. 
\vspace{-12pt}
\subsection{Instruct2Attack (I2A)}
\vspace{-6pt}
Our method aims to solve for $\vect{x}_\text{adv}$ that lies on the natural image manifold, where the difference between $\vect{x}$ and $\vect{x}_\text{adv}$ is controlled by an edit instruction $\vect{L}$. To achieve this, we make use of the learned latent space of a generative model $\mathcal{Z}=\{\vect{z}\ |\ \vect{z}=\mathcal{E}(\vect{x}), \vect{x}\in\mathcal{X}\} \subseteq \mathbb{R}^{h \times w \times c}$, where $\mathcal{E}$ is the image encoder, and $h \times w \times c$ is the dimension of the latent space. Compared to image space, the latent space provides a low-dimensional compact representation that focuses on the important, semantic information of the data, which is more suitable for semantic editing. We then aim to search for an adversarial latent code $\vect{z}_\text{adv}$ according to the conditional distribution $p(\vect{z}|\vect{x}, \vect{L})$, which can be modeled by a pretrained conditional latent diffusion model (LDM)~\cite{brooks2023instructpix2pix, rombach2022high}.

\paragraph{Overview.} The overview of I2A is in~\cref{fig:method} (a). We start with a random noise $\vect{z}_{T}$\,$\backsim$\,$\mathcal{N}(\vect{0}, \sigma_{T}^{2}\vect{I})$ in the latent space, where $\sigma_{T}$ is the noise scale. We then iteratively refine $\vect{z}_{T}$ in the reverse diffusion process using a denoising U-Net~\cite{ronneberger2015u} $\epsilon_\theta$ with the input image $\vect{x}$ and text instruction $\vect{L}$ as conditions to move it towards $p(\vect{z}|\vect{x}, \vect{L})$. At each denoising step, we adversarially guide the diffusion process to reach an adversarial latent code $\vect{z}_\text{adv}$, which is controlled by the learnable adversarial guidance factors $\balpha$ and $\bbeta$ that are obtained by optimizing~\cref{eq:attack_lagrange}. The final output is reconstructed by the image decoder $\mathcal{D}$, \ie, $\vect{x}_{\text{adv}}$\,$=$\,$\mathcal{D}(\vect{z}_\text{adv})$.

\paragraph{Image and text conditioning.} The reverse diffusion process is conditioned on the input image $\vect{x}$ and the edit instruction $\vect{L}$. This is implemented with a conditional denoising U-Net $\epsilon_{\theta}(\vect{z}_{t}, \vect{c}_{I}, \vect{c}_{L})$. Specifically, $\vect{x}$ is first encoded by the image encoder $\mathcal{E}$ to obtain the image condition embedding $\vect{c}_{I}=\mathcal{E}(\vect{x})\in \mathbb{R}^{h \times w \times c}$ and $\vect{L}$ is encoded by a text encoder $\mathcal{C}$ to obtain the text condition embedding $\vect{c}_L=\mathcal{C}(\vect{I})\in \mathbb{R}^{m\times l}$. At each denoising step, $\vect{c}_{I}$ is concatenated with $\vect{z}_t$ as the input to $\epsilon_{\theta}$, while $\vect{c}_{L}$ is projected to the intermediate layers of $\epsilon_{\theta}$ via a cross-attention layer~\cite{vaswani2017attention}.

\paragraph{Adversarial diffusion guidance.} 

Let $\vect{z}_0$ be the denoised output of $\vect{z}_T$ through the reverse diffusion process $\vect{z}_0:=\text{LDM}(\vect{z}_T; \vect{c}_{I}, \vect{c}_{L})$. In a benign editing setting, $\vect{z}_0$ is computed by the following iterative denoising equation~\cite{song2021scorebased}:
\begin{equation}
\scalebox{0.95}{
    $\vect{z}_{t-1} = \vect{z}_{t} + (\sigma_{t}^{2} - \sigma_{t-1}^{2}){e}_{\theta}(\vect{z}_{t}, \vect{c}_{I}, \vect{c}_{L})+ \sqrt{\frac{\sigma_{t-1}^{2}(\sigma_{t}^{2} - \sigma_{t-1}^{2})}{\sigma_{t}^{2}}}\boldsymbol{\zeta}_{t}$,}
    \label{eq:ldm}
\end{equation}
where $1\leq t \leq T$ is the time step, $\sigma_t$ is the noise scale, $\boldsymbol{\zeta}_t\sim \mathcal{N}(\vect{0}, \vect{I})$. ${e}_{\theta}$ is the modified score estimate~\cite{rombach2022high} based on classifier-free guidance~\cite{ho2021classifierfree} to improve sample quality:
\begin{equation}
\begin{split}
    {e}_{\theta}(\vect{z}_t, \vect{c}_I, \vect{c}_L) = &\: \epsilon_{\theta}(\vect{z}_t, \varnothing_I, \varnothing_L) \\ &+ s_I \cdot (\epsilon_{\theta}(\vect{z}_t, \vect{c}_I, \varnothing_L) - \epsilon_{\theta}(\vect{z}_t, \varnothing_I, \varnothing_L)) \\ &+ s_T \cdot (\epsilon_{\theta}(\vect{z}_t, \vect{c}_I, \vect{c}_L) - \epsilon_{\theta}(\vect{z}_t, \vect{c}_I, \varnothing_L)),
    \label{eq:cfg2}
\end{split}
\end{equation}
where the first term is the unconditional guidance, the second term is the image guidance, the third term is the text guidance, $\varnothing_I$ and $\varnothing_L$ are the fixed null values for image and text conditionings respectively, and $s_I$ and $s_T$ are the image and text guidance scales. To adversarially guide the diffusion process to find an adversarial latent code $\vect{z}_\text{adv}$, we introduce two adversarial guidance factors $\balpha, \bbeta\in [0, 1]^{h\times w \times c}$ that adversarially modulate the image guidance and text guidance at each diffusion step, as illustrated in~\cref{fig:method} (b):
\begin{equation}
\begin{split}
    \tilde{e}_{\theta}(\vect{z}_t, \vect{c}_I, \vect{c}_L) & = \: \epsilon_{\theta}(\vect{z}_t, \varnothing_I, \varnothing_L) \\ &+ s_I \cdot \balpha \odot (\epsilon_{\theta}(\vect{z}_t, \vect{c}_I, \varnothing_L) - \epsilon_{\theta}(\vect{z}_t, \varnothing_I, \varnothing_L)) \\ &+ s_T \cdot \bbeta \odot (\epsilon_{\theta}(\vect{z}_t, \vect{c}_I, \vect{c}_L) - \epsilon_{\theta}(\vect{z}_t, \vect{c}_I, \varnothing_L)),
    \label{eq:cfg_adv}
\end{split}
\end{equation}
\noindent where $\odot$ is the Hadamard product. Note that $\balpha$ and $\bbeta$ are not scalars but vectors, which provides more flexibility in controlling the diffusion process. 


\begin{algorithm}[t]
\caption{Instruct2Attack}\label{alg:i2a}
\begin{algorithmic}[1]
\State {\bfseries Input:} $\vect{x}$, $\vect{T}$, $y$, $\gamma$, $\eta$, $\lambda$, $T$, $s_I$, $s_T$, $N$, $f$, $\epsilon_\theta$, $\gE$, $\gD$, $\gC$
\State {\bfseries Output:} adversarial image $\vect{x}_\text{adv}$ 
\LineComment{Initialize the latent code}
\State $\vect{z}_T\sim \mathcal{N}(\vect{0}, \sigma_{T}^{2}\vect{I})$
\LineComment{Image and text encoding}
\State $\vect{c}_I=\gE(\vect{x}),\ \ \vect{c}_L=\gC(\vect{L})$
\LineComment{Optimize $\balpha$ and $\bbeta$} 
\State $\balpha \gets \vec{\mathbf{1}},\ \ \bbeta \gets \vec{\mathbf{1}}$ 
\State {\bfseries for} $i=0$ {\bfseries to} $N-1$
\State $\ \ \ \ \vect{z}_\text{adv}=\text{LDM}(\vect{z}_T; \vect{c}_{I}, \vect{c}_{L}, \balpha, \bbeta$) \Comment{Sample $\vect{z}_\text{adv}$}
\State $\ \ \ \ \vect{x}_{\text{adv}}=\gD(\vect{z}_\text{adv})$ \Comment{Generate $\vect{x}_\text{adv}$}

\State \ \ \ \  $\mathcal{L}=\mathcal{L}_{c}(\vect{x}_{\text{adv}}, y) - \lambda \max\left(0, d(\vect{x}_{\text{adv}}, \vect{x}) -\gamma\right)$
\State \ \ \ \  Update $\balpha$ and $\bbeta$ according to \cref{eq:pgd}
\State \ \ \ \ {\bfseries if} $f(\vect{x}_\text{adv})\neq y$ and $d(\vect{x}_{\text{adv}}, \vect{x})\leq \gamma$ \Comment{Early stopping}
\State \ \ \ \ \ \ \ \ {\bfseries break}
\State \ \ \ \ {\bfseries end if} 
\State {\bfseries end for}
\end{algorithmic}
\end{algorithm}

\paragraph{Perceptual constraint.} To preserve the similarity between $\vect{x}$ and $\vect{x}_\text{adv}$ while allowing for semantic editing, we propose to use the perceptual similarity metric LPIPS~\cite{zhang2018perceptual, laidlaw2021perceptual} instead of the typical pixel-wise $L_p$ distance to bound the adversarial perturbation. The LPIPS distance between two images $\vect{x}_1$ and $\vect{x}_2$ is defined as:
\begin{equation}
    d(\vect{x}_1, \vect{x}_2)\triangleq \lVert \phi(\vect{x}_1)-\phi(\vect{x}_2)\rVert_2,
\end{equation}
where $\phi(\cdot)$ is a feature extraction network. Formally, we aim to solve the following optimization problem: 
\begin{equation}
    \max_{\mathbf{\balpha}, \vect{\bbeta}} \mathcal{L}_{c}(\vect{x}_{\text{adv}}, y),\  \text{s.t.},  d(\vect{x}_{\text{adv}}, \vect{x}) \leq \gamma, 
    \label{eq:attack}
\end{equation} 
where $\vect{x}_{\text{adv}}=\gD(\vect{z}_\text{adv})=\gD(\text{LDM}(\vect{z}_T; \vect{x}, \vect{L}, \balpha, \bbeta))$, $\gamma$ is the perturbation budget, and $\mathcal{L}_{c}$ is the cross-entropy loss. In practice, it is hard to solve Eq.~(\ref{eq:attack}) directly given the LPIPS constraint and we use a Lagrangian relaxation of Eq.~(\ref{eq:attack}): 

\begin{equation}
    \max_{\mathbf{\balpha}, \vect{\bbeta}} \mathcal{L}=\mathcal{L}_{c}(\vect{x}_{\text{adv}}, y) - \lambda \max\left(0, d(\vect{x}_{\text{adv}}, \vect{x}) -\gamma\right), 
    \label{eq:attack_lagrange}
\end{equation} 
where $\lambda$ is the Lagrange multiplier. The second term in~\cref{eq:attack_lagrange} grows linearly with $d(\vect{x}_{\text{adv}}, \vect{x})$ when it is greater than $\gamma$ and becomes 0 when the perceptual constraint $d(\vect{x}_{\text{adv}}, \vect{x})\leq\gamma$ is satisfied. We iteratively solve~\cref{eq:attack_lagrange} with projected gradient descent:

\vspace{-10pt}
\begin{numcases}{}
    {\balpha}^{(i+1)} = \gP \left({\balpha}^{(i)} + \eta \cdot \text{sign} \nabla_{\vect{\alpha}^{(i)}} \mathcal{L}\right), \nonumber \\
    \bbeta^{(i+1)} = \gP \left(\bbeta^{(i)} + \eta \cdot \text{sign} \nabla_{\bbeta^{(i)}} \mathcal{L}\right), 
    \label{eq:pgd}
\end{numcases}
where $\gP$ is the projection onto $[0, 1]^{h\times w \times c}$, $0\leq i <N$ is the attack step, $N$ is the maximum attack iteration, $\eta$ is the step size, and $\balpha$ and $\bbeta$ are initialized as all-one vectors $\vec{\mathbf{1}}$.
The I2A algorithm is summarized in~\cref{alg:i2a}. At the end of the attack generation, we ensure the perceptual constraint is satisfied by projecting $\vect{x}_{\text{adv}}$ back to the feasible set through dynamically adjusting $s_I$ and $s_T$ in~\cref{eq:cfg_adv} which control the trade-off between how strongly the edited image corresponds with the input image and the edit instruction. The details of the projection algorithm are in the Supp. 

\begin{figure}[t]
\centering
\small
\setlength{\tabcolsep}{0.8mm}
\scalebox{0.93}{
\begin{tabular}[b]{c p{0.142\textwidth}p{0.142\textwidth}p{0.142\textwidth}}    
       & 
    \begin{subfigure}{0.142\textwidth}
        \includegraphics[width=\textwidth]{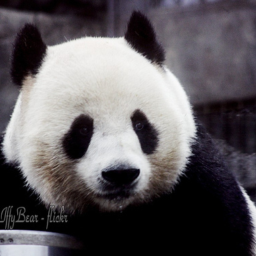}
    \end{subfigure} & 
    \begin{subfigure}{0.142\textwidth}
        \includegraphics[width=\textwidth]{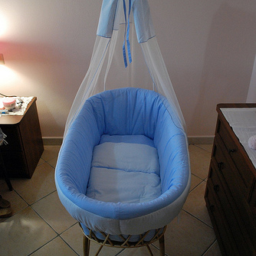}
    \end{subfigure} & 
    \begin{subfigure}{0.142\textwidth}
        \includegraphics[width=\textwidth]
        {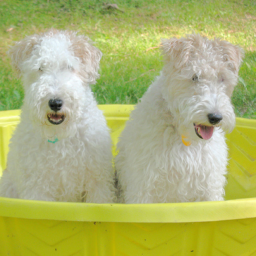}
    \end{subfigure} \\ \toprule
    \textbf{BLIP-2} & \textit{``a panda bear is looking at the camera"} & \textit{``a baby's crib with a blue canopy and a white net" }& \textit{``two white dogs sitting in a yellow tub"}\\ \midrule
    \textbf{GPT-4} & \colorbox{second}{\parbox{0.13\textwidth}{\textit{``Add bamboo in background."}}} & \colorbox{second}{\parbox{0.13\textwidth}{\textit{``Add a teddy bear."}}} & \colorbox{second}{\parbox{0.13\textwidth}{\textit{``Fill the tub with water."}}}\\
\end{tabular}} 
\vspace{-7pt}
\caption{\textbf{Examples of automatic instruction generation.} We first use BLIP-2~\cite{li2023blip} to generate image captions and feed them into GPT-4~\cite{openai2023gpt} to generate image-specific semantic edit instructions.}
\vspace{-10pt}
\label{fig:gpt4}
\end{figure}


\subsection{Automatic Instruction Generation}
\label{sec:gpt4}
An important component of I2A is the input text instruction. While the text instructions can be any user-defined prompts, they should be legitimate to produce natural and plausible semantic modifications without alternating object categories. Ideally, the text instruction should be determined based on the image context. However, it is infeasible to manually write such image-specific prompts for every image in a large dataset. To address this challenge, we leverage large vision and language models to automatically generate image-specific editing instructions. We first convert the input image to an image caption using BLIP-2~\cite{li2023blip}. Then we apply the language-only GPT-4~\cite{openai2023gpt} to generate editing instructions based on the input captions and object categories. We utilize few-shot in-context learning by presenting 50 manually written examples in the prompt without fine-tuning the GPT-4 model. GPT-4 is instructed to \textit{``generate natural and simple image editing instructions without altering the inherent nature or category of objects within the image"}. The full prompt for GPT-4 can be found in the supplementary material. Examples of automatically generated image captions and edit instructions are shown in~\cref{fig:gpt4}. The proposed method can generate relevant image editing instructions based on the image content.

\begin{table*}[t]
\vspace{-10pt}
\centering
\setlength{\tabcolsep}{1.4mm}
\scalebox{0.9}{\begin{tabular}{c | l | c c c c c c c c c c | c} 
\toprule
Model & Method & Clean & FGSM & PGD & MIM & Auto & Fog & Gabor & Snow & StAdv & PerC-AL  & \textbf{I2A (Ours)} \\
\midrule
\multirow{3}{*}{\rotatebox{0}{ResNet-18}} & Undefended &69.68        & 0.90  & {0.00}  & {0.00}  & 0.00      & 5.24  & 0.20    & 0.26  & 0.00 & 56.16   & 1.20 \\
 & Salman2020~\cite{salman2020adversarially} & 52.92        & 32.54 & 29.77 & 29.84 & 25.32      & 27.14 & 33.44  & 30.46 & 29.40 & \second{9.92}    & \best{0.54} \\
  & DiffPure~\cite{nie2022DiffPure} & 64.70        & 58.98 & 62.16 & 61.36 & 63.22 &  51.20 & 59.84 & 53.34 & \second{49.58} & 64.74 & \best{42.47}\\
\midrule
\multirow{5}{*}{\rotatebox{0}{ResNet-50}} & Undefended & 76.52        & 5.96  & 0.00 & 0.00 & 0.00      & 6.76  & {0.04}      & 1.12  & 0.00 & 72.34   & 2.11 \\
 & Salman2020~\cite{salman2020adversarially} & 64.02        & 43.44 & 38.98 & 39.06 & 34.96      & 36.22 & 41.44 & 40.36    & 36.56 & \second{18.68}   & \best{1.42}   \\
 & Engstrom2019~\cite{engstrom2019learning}  & 62.56        & 39.92 & 32.92 & 33.18 & 29.22      & 37.30 & 37.24   & 37.94 & 36.46 & \second{20.96}   & \best{2.52}  \\
 & FastAT~\cite{Wong2020Fast} & 55.62        & 32.84 & 27.94 & 26.26 & 26.24      & 19.54 & 30.76  & 27.28 & 31.90 & \second{13.74}   & \best{1.29}  \\
    & DiffPure~\cite{nie2022DiffPure} & 70.54        & 66.94 & 69.16 & 68.06 & 69.68 &  58.90 & 66.42  & 61.62 & \second{56.20}  & 70.48 & \best{46.87}\\
\midrule
\multirow{4}{*}{\rotatebox{0}{ConvNeXt-B}} & Undefended & 83.36        & 41.44 & 0.00  & 0.00  & 0.00       & 11.16 & 0.14   & 6.86  & 0.70  & 19.94   & 8.93  \\
 & ARES~\cite{liu2023comprehensive} & 76.02        & 59.60 & 56.28 & 56.60 & 55.82      & 59.41 & 58.06  & 59.36 & \second{47.56} & 74.66   & \best{4.12}  \\
 & ConvStem~\cite{singh2023revisiting} & 75.90        & 59.92 & 56.74 & 56.88 & 56.14      & 60.12 & 58.64  & 59.32 & \second{49.56} & 75.22   & \best{3.62} \\
    & DiffPure~\cite{nie2022DiffPure}& 78.10        & 74.90 & 77.32 & 76.58 & 77.54 &  69.28 & 75.04 & \second{54.82} & 70.92 & 75.74 & \best{53.34}  \\
\midrule
\multirow{4}{*}{\rotatebox{0}{ConvNeXt-L}} & Undefended & 83.62        & 44.44 & 0.00  & 0.00  & 0.00       & 9.38  & 0.18   & 6.72  & 1.60  & 16.08   & 9.32  \\
 & ARES~\cite{liu2023comprehensive} & 78.02        & 62.00 & 58.60 & 58.76 & 58.48      & 60.52 & 60.78 &  61.74 & \second{51.22} & 77.02   & \best{4.95}\\
 & ConvStem~\cite{singh2023revisiting} & 77.00        & 60.58 & 57.70 & 58.00 & 57.70      & 60.28 & 59.96 & 60.54 & \second{49.22} & 76.62   & \best{4.91}\\
    & DiffPure~\cite{nie2022DiffPure}& 78.68        & 75.12 & 77.84 & 77.02 &     78.12 &  \second{69.84}	& 75.32		& 73.66 & 71.52 & 78.12 & \best{52.71} \\
\midrule
\multirow{3}{*}{\rotatebox{0}{Swin-B}} & Undefended & 81.98        & 43.50 & 0.00  & 0.00  & 0.00       & 12.24 & 0.30 & 6.64  & 2.52  & 51.02   & 8.81  \\
 & ARES~\cite{liu2023comprehensive} & 76.16        & 59.78 & 57.12 & 57.32 & 56.16      & 59.18 & 58.76  & 58.90 & 47.76 & \second{41.06}   & \best{3.47}\\
    & DiffPure~\cite{nie2022DiffPure} & 76.60        & 74.02 & 75.88 & 75.28 &       75.92 & \second{68.64}	& 73.50	& 70.90  & 70.12 &  76.54 & \best{50.65}\\
\midrule
\multirow{3}{*}{\rotatebox{0}{Swin-L}} & Undefended & 85.06        & 45.72 & 0.00  & 0.00  & 0.00       & 6.92  & 0.78  & 8.72  & 2.96  & 57.62   & 8.40 \\
 & ARES~\cite{liu2023comprehensive} & 78.92        & 62.84 & 59.44 & 59.72 & 59.56      & 60.82 & 61.60    & 62.34 & \second{51.62} & 62.56   & \best{4.98}  \\
    & DiffPure~\cite{nie2022DiffPure} & 81.40         & 79.56 & 80.66 & 80.26 &  81.02 &   \second{73.56}	& 78.72	& 75.9  & 74.48 &  80.92 & \best{55.18}\\
\midrule
\multicolumn{2}{c|}{Average Accuracy} & 73.97 & 51.13 & 41.75 & 41.55 & 41.14 & 41.98 & 42.33 & 41.76 & \second{37.81} & 54.00 & \best{16.90}\\ \bottomrule
\end{tabular}}
\vspace{-5pt}
\caption{\textbf{White-box attacks.} We report top-1 (\%) on ImageNet. The best for defended models are in \best{bold} and the 2nd best are \second{{underlined}}. }
\label{tab:wb}
\vspace{-10pt}
\end{table*}

\section{Experiments and Results}
\subsection{Settings}
\label{sec:settings}
\looseness -1 \paragraph{Datasets and metrics.} We evaluate I2A on the ImageNet~\cite{deng2009imagenet} subset used by Robustbench~\cite{croce2021robustbench}, which consists of 5,000 randomly sampled images from the ImageNet validation set. We additionally evaluate I2A on Places365~\cite{zhou2017places}, a scene recognition dataset with 365 scene classes. We randomly selected ten images for each class from the validation set, resulting in a subset of 3,650 images. All images are resized to $256\times 256\times 3$. We adopt top-1 classification accuracy to evaluate the attack effectiveness. 
\paragraph{Implementation details.} We use four manually written editing instructions and the instructions generated by GPT4 for evaluation, resulting in five editing prompts per image. Unless mentioned otherwise, we report the average accuracy of the five edit prompts for I2A. The manually written instructions include instructions that change the lighting (\texttt{"make it at night"}), weather conditions (\texttt{"make it in snow"}), and overall image styles (\texttt{"make it a sketch painting"}, \texttt{"make it a vintage photo"}). These are general edits applicable to every image. We set $\lambda$\,$=$\,$100$, $\gamma$\,$=$\,$0.3$, $\eta$\,$=$\,$0.1$, $T$\,$=$\,$20$, $s_I$\,$=$\,$1.5$, $s_T$\,$=$\,$7.5$, and $N$\,$=$\,$200$. The latent space dimension is $32$\,$\times$\,$32$\,$\times$\,$4$. We use the pretrained AlexNet~\cite{krizhevsky2012imagenet} as the feature network $\phi$ for LPIPS. For $\mathcal{E}$, $\mathcal{D}$, $\mathcal{C}$ and $\epsilon_\theta$, we use the pretrained weights of~\cite{brooks2023instructpix2pix} and keep the weights frozen during attack generation. More details are in the Supp.


\subsection{Results on ImageNet}
\label{sec:imgnet}
\paragraph{Baselines.} We consider $L_p$ attacks including FGSM~\cite{goodfellow2014explaining}, PGD~\cite{madry2018towards}, MIM~\cite{dong2018boosting} and AutoAttack~\cite{croce2020reliable}, and non-$L_p$ attacks including Fog, Gabor, Elastic, Snow~\cite{kang2019testing}, StAdv~\cite{xiao2018spatially} and PerC-AL~\cite{zhao2020towards} for comparisons. For $L_p$ attacks, we use $L_{\infty}$ norm with a bound of $4/255$. The implementation details of the baselines can be found in the Supp.

\paragraph{Victim models.} We consider a variety of networks, from convolutional ResNet-18/50~\cite{he2016deep} and the recent ConvNeXt-B/L~\cite{liu2022convnet}, to Transformer-based Swin-B/L~\cite{liu2021swin}. 
\paragraph{Defense methods.} To better understand the attack performance under defenses, we evaluate both \textit{undefended} and \textit{defended} models for each network architecture. We consider adversarial training-based defenses, including models trained by Salman2020~\cite{salman2020adversarially}, Engstrom2019~\cite{engstrom2019learning}, Fast AT~\cite{Wong2020Fast}, ARES-Bench~\cite{liu2023comprehensive}, and ConvStem~\cite{singh2023revisiting}, which are top-performing models on the Robustbench ImageNet leaderboard~\cite{croce2021robustbench}. We also evaluate the attack methods on a state-of-the-art preprocessing-based defense DiffPure~\cite{nie2022DiffPure}, which aims to remove adversarial perturbations using a diffusion model. Details of defense methods are in the Supp.

\paragraph{White-box attack.} We compare I2A with baseline attacks under the white-box attack setting in~\Cref{tab:wb}. Overall, I2A achieves the lowest average classification accuracy, which is significantly lower than the second best method StAdv~\cite{xiao2018spatially} (16.90\% \textit{vs.} 37.81\%). In addition, we find that I2A is highly effective for adversarially trained models. In particular, Swin-L trained by ARES~\cite{liu2023comprehensive} is the current strongest model on the Robustbench ImageNet leaderboard~\cite{croce2021robustbench}, which achieves 59.56\% accuracy under AutoAttack~\cite{croce2020reliable} but only 4.98\% under the proposed I2A attack. These results suggest that I2A reveals new vulnerabilities in DNNs. Another interesting finding from~\Cref{tab:wb} is that undefended models are slightly more robust to the I2A attack than adversarially trained models. For example, the undefended ConvNeXt-L model achieves 9.32\% accuracy under I2A attack while the adversarially trained model from~\cite{liu2023comprehensive} only achieves 4.95\% accuracy. However, this phenomenon is not observed for the baseline attacks. We believe this is because the adversarial examples generated by I2A lie on the natural image manifold, while adversarial training hurts the model performance on natural images~\cite{raghunathan2020understanding} as indicated by the degraded performance on clean images. 

I2A also remains much more effective under the strong preprocessing defense DiffPure~\cite{nie2022DiffPure} compared to the baseline attacks since I2A generates natural semantic perturbations that are less likely to be removed in the denoising diffusion process. For example, for the Swin-L model, I2A achieves 55.18\% classification accuracy under the DiffPure defense, while the second best method Fog~\cite{kang2019testing} achieves 73.56\% accuracy, which is 18.38\% higher than I2A. Note that the results of DiffPure in~\Cref{tab:wb} are under non-adaptive attacks where the attacker does not attempt to attack the defense mechanism. We further investigate the adaptive attack scenario, where the attacker uses the BPDA+EOT techniques~\cite{athalye2018obfuscated, tramer2020adaptive} to attack the DiffPure defense.  Due to the high computational cost, we experiment with 200 images, and for I2A attack we use the prompt \texttt{"make it in snow"}. The results are summarized in~\Cref{tab:bpda}. I2A can successfully break the DiffPure defense under the adaptive attack setting, bringing down the accuracy to as low as 7.0\%. However, DiffPure demonstrates strong robustness the noise-based PGD~\cite{madry2018towards} even under strong adaptive attacks, \eg, achieving 68.5\% accuracy for the Swin-B model. StAdv~\cite{xiao2018spatially} is a semantic attack based on local geometric transformations, which is more effective than PGD in attacking DiffPure, but still performs worse than I2A.

\begin{table}
\centering
\setlength{\tabcolsep}{2.0mm}
\scalebox{0.9}{\begin{tabular}{c|c| c c c c} 
\toprule
Model & Adaptive & Clean & PGD & StAdv & \textbf{I2A (Ours)} \\
\midrule
\multirow{2}{*}{ResNet-18} & \ding{55} & 70.0 & 68.5 & \underline{57.0} & \textbf{43.5} \\
 & \ding{51} & 70.0 & 44.0 & \underline{34.0} & \textbf{7.0 }\\
 \midrule
\multirow{2}{*}{ResNet-50} & \ding{55} & 75.0 & 75.0 & \underline{59.0} & \textbf{52.5} \\
 & \ding{51} &  75.0 & 56.0 & \underline{40.5} &  \textbf{9.0}\\
 \midrule
\multirow{2}{*}{Swin-B} & \ding{55} & 80.0 & 75.5 & \underline{69.0} & \textbf{52.0} \\
 & \ding{51} & 80.0 & 68.5 & \underline{52.5} &  \textbf{20.5}\\
\bottomrule
\end{tabular}}
\vspace{-7pt}
\caption{\textbf{DiffPure defense under adaptive and non-adaptive attacks}~\cite{nie2022DiffPure}. We report top-1 (\%) on ImageNet. }
\vspace{-10pt}
\label{tab:bpda}
\end{table}

\begin{figure*}[t]
\centering
\hspace*{-0.6cm}  
\small
\setlength{\tabcolsep}{0mm}
\scalebox{1.0}{
\begin{tabular}[b]{ cccc}
    \begin{subfigure}{0.25\textwidth}
        \includegraphics[width=\textwidth]{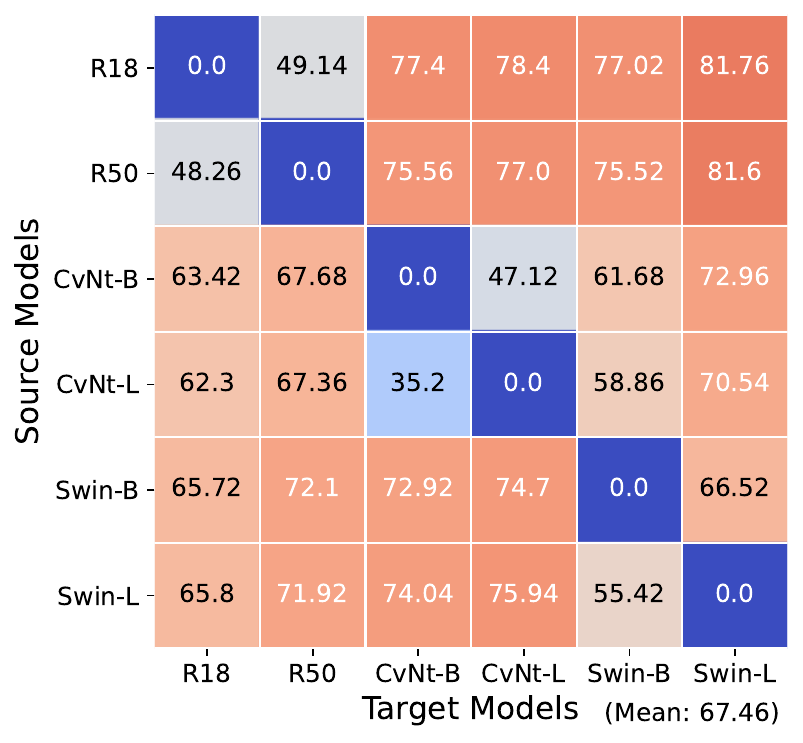}
    \end{subfigure} & 
    \begin{subfigure}{0.25\textwidth}
        \includegraphics[width=\textwidth]{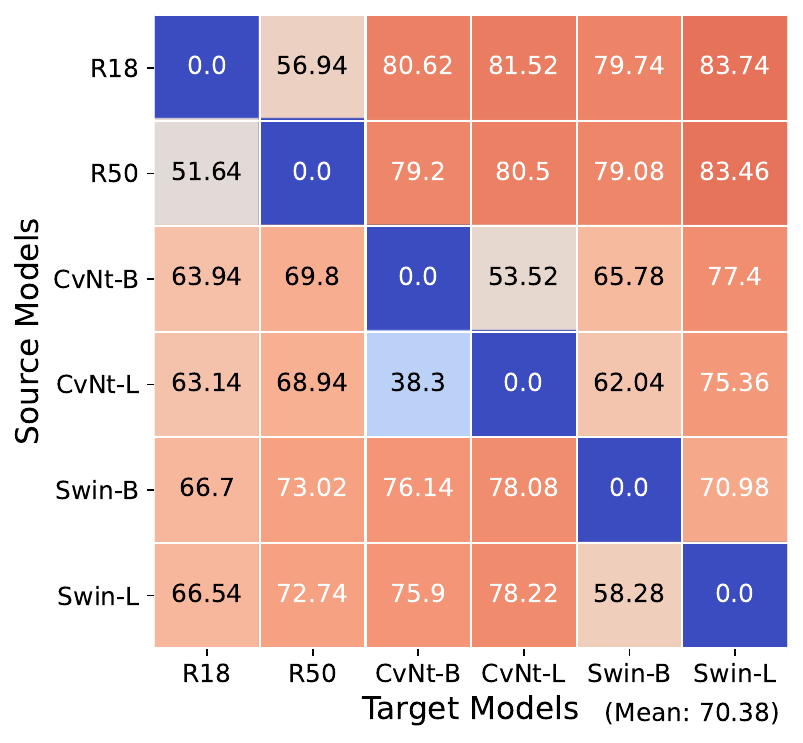}
    \end{subfigure} & 
    \begin{subfigure}{0.25\textwidth}
        \includegraphics[width=\textwidth]
        {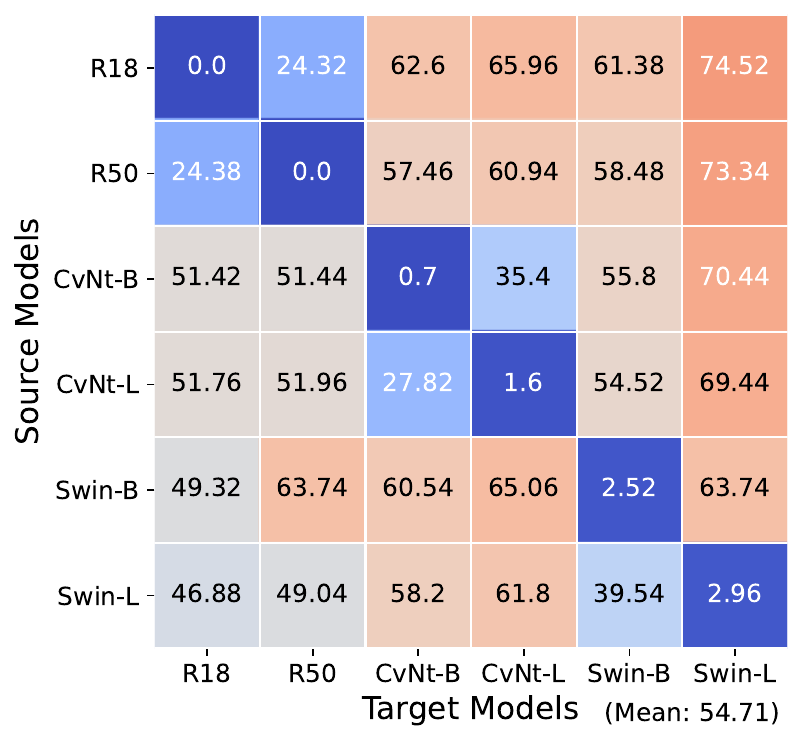}
    \end{subfigure} & 
    \begin{subfigure}{0.293\textwidth}
        \includegraphics[width=\textwidth]{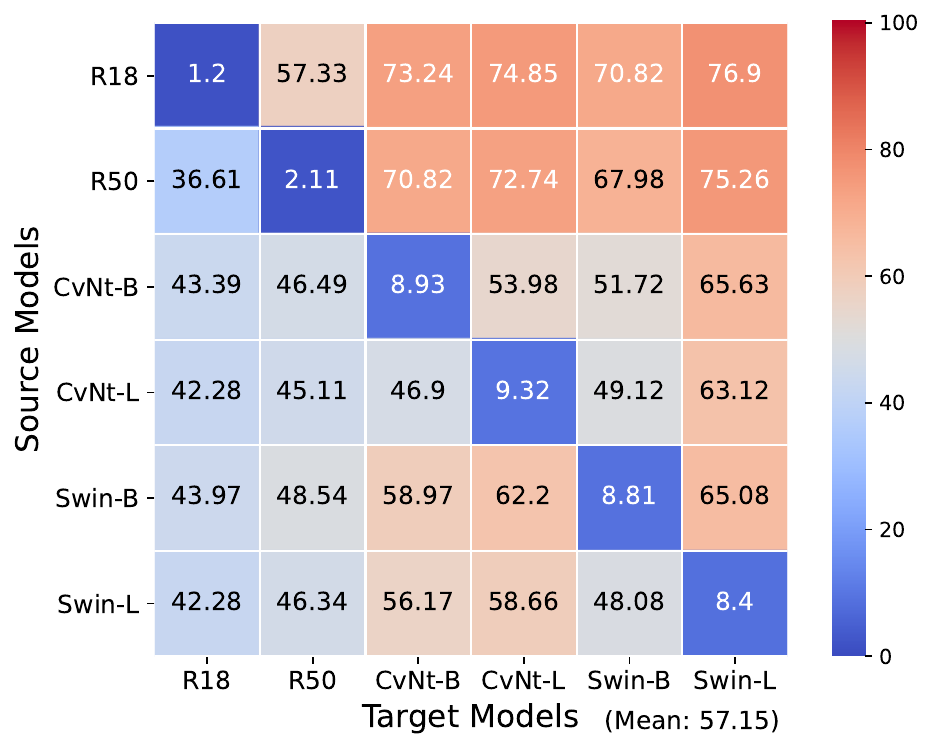}
    \end{subfigure}  \\
    \begin{subfigure}{0.25\textwidth}
        \includegraphics[width=\textwidth]{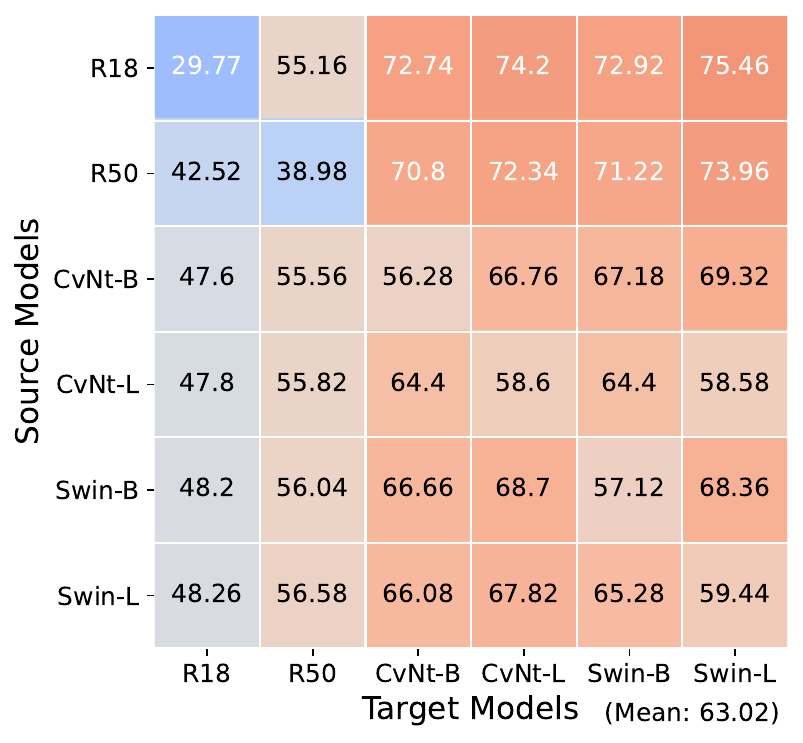}
        \caption{PGD~\cite{madry2018towards}}
    \end{subfigure} & 
    \begin{subfigure}{0.25\textwidth}
        \includegraphics[width=\textwidth]{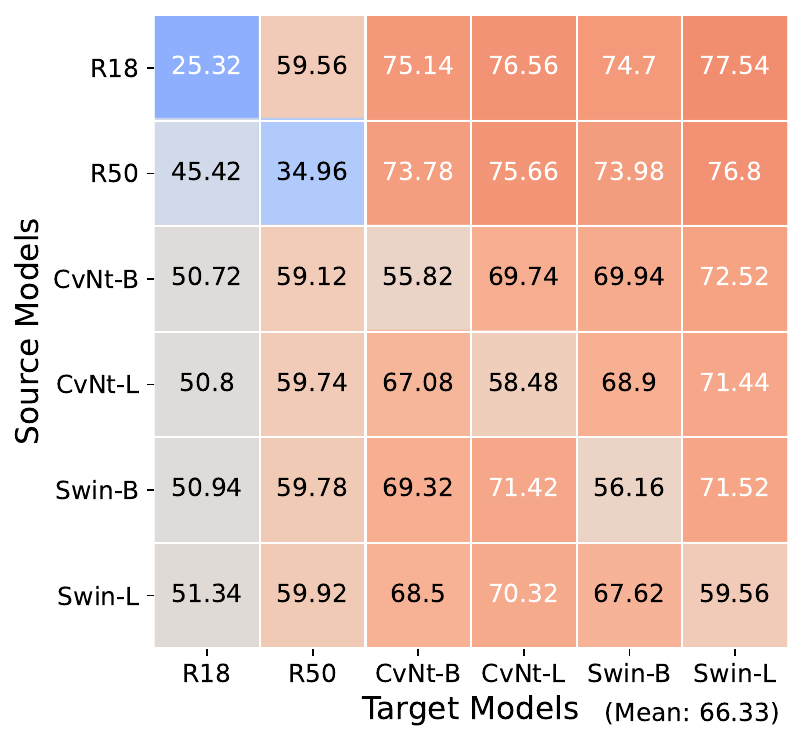}
        \caption{AutoAttack~\cite{croce2020reliable}}
    \end{subfigure} & 
    \begin{subfigure}{0.25\textwidth}
        \includegraphics[width=\textwidth]
        {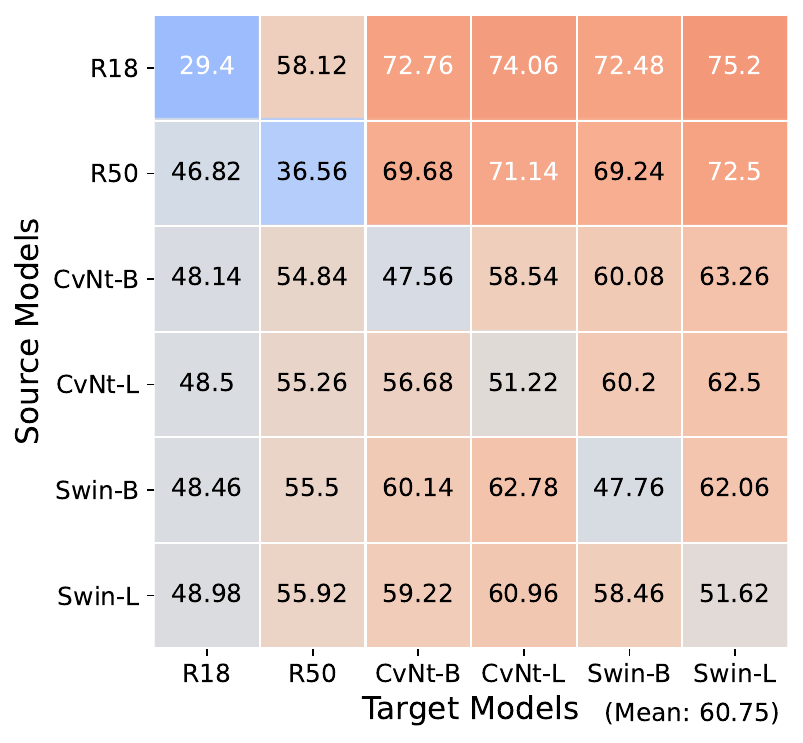}
        \caption{StAdv~\cite{xiao2018spatially}}
    \end{subfigure} & 
    \begin{subfigure}{0.293\textwidth}
        \includegraphics[width=\textwidth]{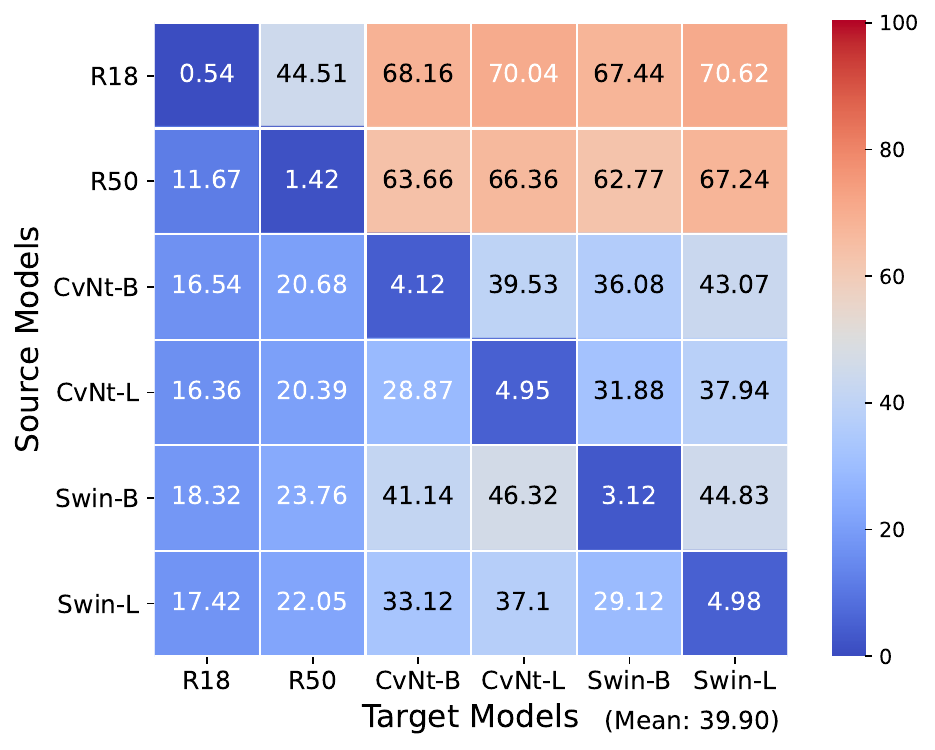}
        \caption{\textbf{I2A (Ours)}}
    \end{subfigure}  \\
\end{tabular}}
\vspace{-20pt}
\caption{\textbf{Black-box attacks.} We report top-1 (\%) on ImageNet. \textit{Top}: transferability between undefended models. \textit{Bottom}: transferability between adversarially trained models. Average accuracy of black-box attacks is marked on the bottom right of each figure.}
\vspace{-7pt}
\label{fig:blackbox}
\end{figure*}

\paragraph{Black-box attack.} We also consider the black-box attack scenarios where we craft the attacks on a source model and evaluate them on the target models. The results are presented in~\cref{fig:blackbox}, where we show the transferability within undefended models as well as adversarially trained models. For adversarially trained models, we use models trained by~\cite{salman2020adversarially} for ResNet-18 and ResNet-50, and models trained by~\cite{liu2023comprehensive} for the other network architectures. In general, attacks crafted on stronger models such as Swin-B and Swin-L tend to transfer better to weaker models such as ResNet-18 and ResNet-50, indicated by lower values in the lower triangles than the upper triangles of the figures. In addition, the attacks seem to transfer better within adversarially trained models than undefended models, resulting in lower classification accuracy. Overall, semantic attacks such as I2A and StAdv~\cite{xiao2018spatially} achieve better transferability than noise-based attacks such as PGD~\cite{madry2018towards} and AutoAttack~\cite{croce2020reliable}. For undefended models, I2A achieves slightly higher average classification accuracy than StAdv (57.15\% \textit{vs.} 54.71\%). However, I2A obtains much lower average classification accuracy than StAdv for adversarially trained models (39.90\% \textit{vs.} 60.75\%), indicating better transferability of I2A. In particular, the I2A attack crafted on Swin-L achieves 17.42\% accuracy when evaluated on ResNet-18, resulting in a black-box attack success rate of \textbf{82.58\%}. The great transferability of I2A suggests that it captures the common vulnerabilities among different models.

\begin{figure*}[t]
\centering
\small
\setlength{\tabcolsep}{1mm}
\scalebox{0.892}{
\begin{tabular}[b]{ cccccc}
    \begin{subfigure}{0.16\textwidth}
        \includegraphics[width=\textwidth]{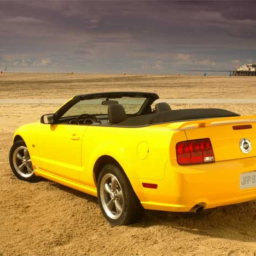}
        \caption*{Clean Image}
    \end{subfigure} & 
    \begin{subfigure}{0.16\textwidth}
        \includegraphics[width=\textwidth]{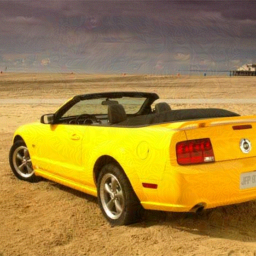}
        \caption*{PGD~\cite{madry2018towards}}
    \end{subfigure} & 
    \begin{subfigure}{0.16\textwidth}
        \includegraphics[width=\textwidth]{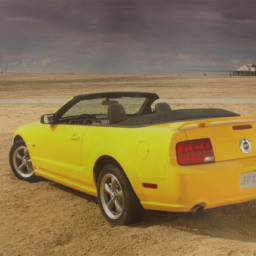}
        \caption*{Fog~\cite{kang2019testing}}
    \end{subfigure} & 
    \begin{subfigure}{0.16\textwidth}
        \includegraphics[width=\textwidth]{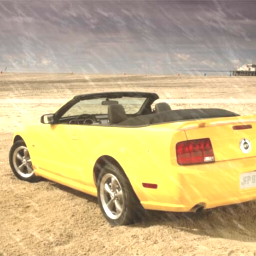}
        \caption*{Snow~\cite{kang2019testing}}
    \end{subfigure} & 
    \begin{subfigure}{0.16\textwidth}
        \includegraphics[width=\textwidth]{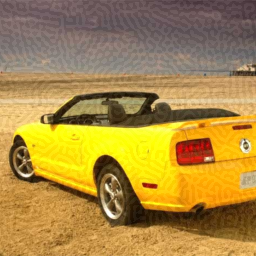}
        \caption*{Gabor~\cite{kang2019testing}}
    \end{subfigure} & 
    \begin{subfigure}{0.16\textwidth}
        \includegraphics[width=\textwidth]{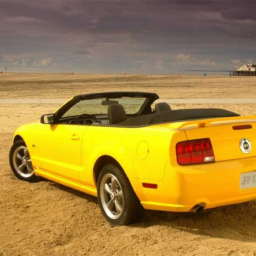}
        \caption*{PerC-AL~\cite{zhang2018perceptual}}
    \end{subfigure} \\
    \begin{subfigure}[t]{0.16\textwidth}
        \includegraphics[width=\textwidth]{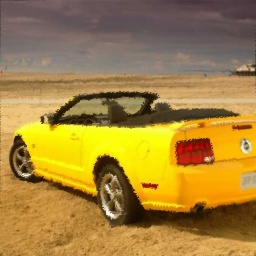}
        \caption*{StAdv~\cite{xiao2018spatially}}
    \end{subfigure} & 
    \begin{subfigure}[t]{0.16\textwidth}
        \includegraphics[width=\textwidth]{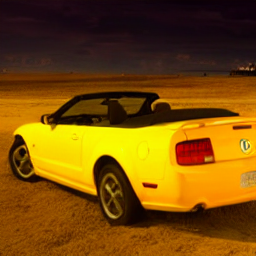}
        \caption*{\textbf{I2A:} \textit{``make it at night"}}
    \end{subfigure} & 
    \begin{subfigure}[t]{0.16\textwidth}
        \includegraphics[width=\textwidth]{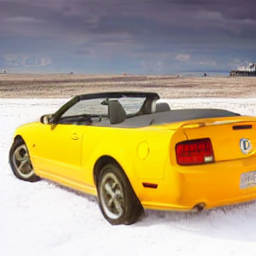}
        \caption*{\textbf{I2A:} \textit{``make it in snow"}}
    \end{subfigure} & 
    \begin{subfigure}[t]{0.16\textwidth}
        \includegraphics[width=\textwidth]{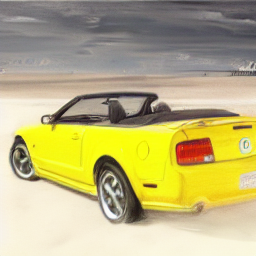}
        \caption*{\textbf{I2A:} \textit{``make it a sketch painting"}}
    \end{subfigure} & 
    \begin{subfigure}[t]{0.16\textwidth}
        \includegraphics[width=\textwidth]{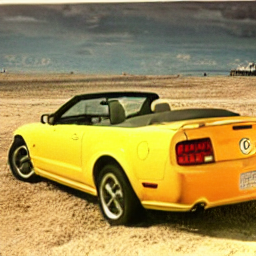}
        \caption*{\textbf{I2A:} \textit{``make it a vintage photo"}}
    \end{subfigure} & 
    \begin{subfigure}[t]{0.16\textwidth}
        \includegraphics[width=\textwidth]{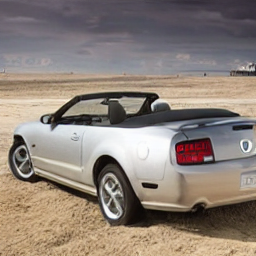}
        \caption*{\textbf{I2A:} \textit{``make it silver"} (GPT-4)}
     \end{subfigure}  \\ \hline 
    \vspace{-5pt}\\
    \begin{subfigure}[b]{0.16\textwidth}
        \includegraphics[width=\textwidth]{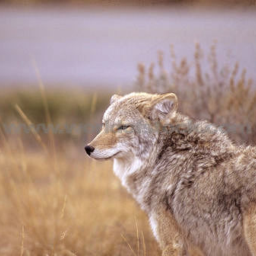}
        \caption*{Clean Image}
    \end{subfigure} & 
    \begin{subfigure}[b]{0.16\textwidth}
        \includegraphics[width=\textwidth]{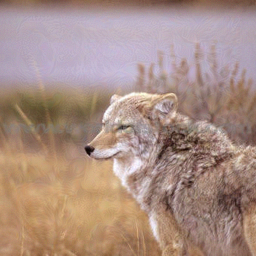}
        \caption*{PGD~\cite{madry2018towards}}
    \end{subfigure} & 
    \begin{subfigure}[b]{0.16\textwidth}
        \includegraphics[width=\textwidth]{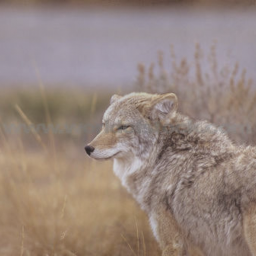}
        \caption*{Fog~\cite{kang2019testing}}
    \end{subfigure} & 
    \begin{subfigure}[b]{0.16\textwidth}
        \includegraphics[width=\textwidth]{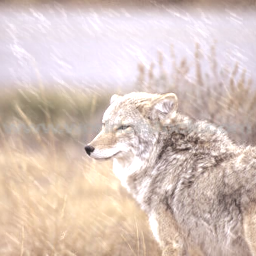}
        \caption*{Snow~\cite{kang2019testing}}
    \end{subfigure} & 
    \begin{subfigure}[b]{0.16\textwidth}
        \includegraphics[width=\textwidth]{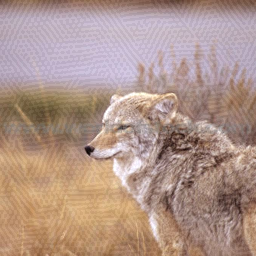}
        \caption*{Gabor~\cite{kang2019testing}}
    \end{subfigure} & 
    \begin{subfigure}[b]{0.16\textwidth}
        \includegraphics[width=\textwidth]{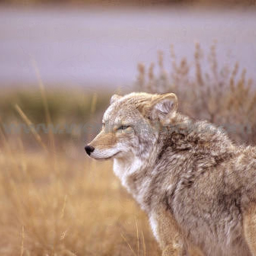}
        \caption*{PerC-AL~\cite{zhao2020towards}}
    \end{subfigure} \\
    \begin{subfigure}[t]{0.16\textwidth}
        \includegraphics[width=\textwidth]{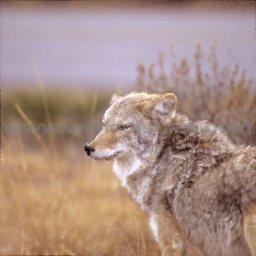}
        \caption*{StAdv~\cite{xiao2018spatially}}
    \end{subfigure} & 
    \begin{subfigure}[t]{0.16\textwidth}
        \includegraphics[width=\textwidth]{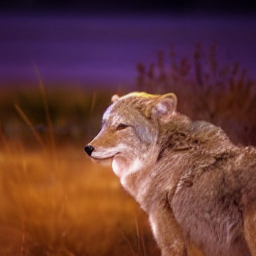}
        \caption*{\textbf{I2A:} \textit{``make it at night"}}
    \end{subfigure} & 
    \begin{subfigure}[t]{0.16\textwidth}
        \includegraphics[width=\textwidth]{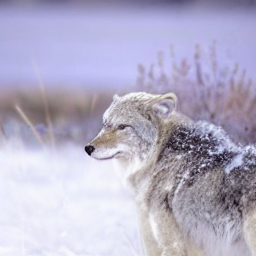}
        \caption*{\textbf{I2A:} \textit{``make it in snow"}}
    \end{subfigure} & 
    \begin{subfigure}[t]{0.16\textwidth}
        \includegraphics[width=\textwidth]{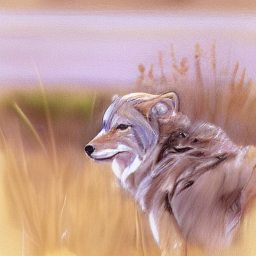}
        \caption*{\textbf{I2A:} \textit{``make it a sketch painting"}}
    \end{subfigure} & 
    \begin{subfigure}[t]{0.16\textwidth}
        \includegraphics[width=\textwidth]{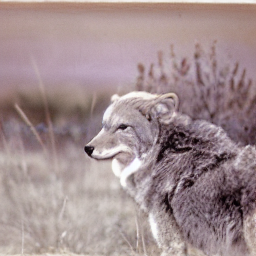}
        \caption*{\textbf{I2A:} \textit{``make it a vintage photo"}}
    \end{subfigure} & 
    \begin{subfigure}[t]{0.16\textwidth}
        \includegraphics[width=\textwidth]{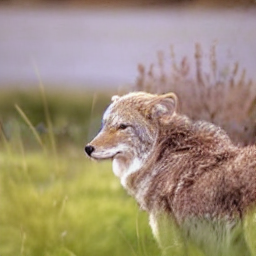}
        \caption*{\textbf{I2A:} \textit{``make the field green"} (GPT-4)}
    \end{subfigure} \\
\end{tabular}} 
\vspace{-9pt}
\caption{\textbf{Visualization of adversarial images.} I2A generates natural and diverse perturbations based on the text instructions.}
\vspace{-8pt}
\label{fig:vis}
\end{figure*}

\paragraph{Visualization.} 
In~\cref{fig:vis}, we present adversarial images produced by the various attack methods. PGD~\cite{madry2018towards} generates adversarial images by adding subtle adversarial noise, which creates unnatural adversarial patterns on the image. Fog and Snow attacks~\cite{kang2019testing} introduce adversarially chosen partial occlusion of the image to mimic the effect of mist and snowflakes, yet these perturbations appear artificial. Gabor attack~\cite{kang2019testing} superimposes adversarial Gabor noise onto the image, which is conspicuous and compromises image quality. Perc-AL~\cite{zhao2020towards} minimizes perturbation sizes with respect to perceptual color distance, resulting in large yet imperceptible perturbations, but the attack strength is relatively weak as shown in~\Cref{tab:wb}. StAdv attack~\cite{xiao2018spatially} introduces adversarial local geometric transformations which distort the input image. The proposed I2A attack generates natural and diverse perturbations based on the text instructions, resulting in visually appealing adversarial images.
More visualizations can be found in the Supp. 

\subsection{Results on Places365}

\begin{table}
\centering
\setlength{\tabcolsep}{1.5mm}
\scalebox{0.9}{\begin{tabular}{c | l | c c c c c} 
\toprule
Model & Method & Clean & PGD & StAdv & \textbf{I2A} \\
\midrule
\multirow{2}{*}{ResNet-18} & Undefended & 52.96 & 0.0 & 0.17 & 0.41 \\
 & DiffPure~\cite{nie2022DiffPure} & 50.88 & 44.63 & \second{36.26} & \best{26.68} \\
\midrule
\multirow{2}{*}{ResNet-50} & Undefended & 53.70 & 0.0 & 0.10 &  0.61 \\
 & DiffPure~\cite{nie2022DiffPure} & 52.25 & 47.86 & \second{39.51} & \best{29.06} \\
\midrule
\multirow{2}{*}{DenseNet-161} & Undefended & 54.50 & 0.0 &  0.27 & 0.73 \\
 & DiffPure~\cite{nie2022DiffPure} & 51.69 & 48.75 & \second{40.31} & \best{30.71} \\
\midrule
\multicolumn{2}{c|}{Average Accuracy} & 52.66 & 23.54 & \second{19.44} & \best{14.70} \\
\bottomrule
\end{tabular}}
\vspace{-7pt}
\caption{\textbf{White-box attacks on Places365}. We report top-1 (\%). The best for DiffPure are in \best{bold} and the 2nd best are \second{{underlined}}.}
\vspace{-10pt}
\label{tab:places365}
\end{table}

To demonstrate the generalizability of our method to different data domains, we also evaluate I2A on Places365~\cite{zhou2017places}. The results are shown in \Cref{tab:places365}. We compare I2A to PGD~\cite{madry2018towards} and StAdv~\cite{qiu2020semanticadv} attacks. All three methods can break the undefended models, bringing down the classification accuracy to almost zero. However, I2A achieves much lower classification accuracy when the DiffPure defense is used. Visualizations of Places365 can be found in the Supp.

\subsection{Ablation Studies}
We conduct ablations on a subset of 500 images on the undefended models. We use the prompt \texttt{"make it in snow"}. The other settings are the same as those in~\cref{sec:settings}.
\paragraph{Effect of adversarial diffusion guidance.} 
We demonstrate the effect of the adversarial guidance factors $\balpha$ and $\bbeta$ of~\cref{eq:cfg_adv} in~\Cref{tab:ab}. When we do not apply $\balpha$ and $\bbeta$ (row 1), which is the benign editing setting, the classification accuracies are similar to these on clean images. When we introduce $\bbeta$ to the text guidance term (row 2) or $\balpha$ to the image guidance term (row 3), the classification accuracies drop significantly. In addition, it can be observed that modulating the image guidance term is more effective than the text guidance term by comparing row 2 with row 3. The attacks are most effective when we apply both $\balpha$ and $\bbeta$. These results demonstrate the effectiveness of the proposed adversarial diffusion guidance method.

\begin{table}
    \centering
    \setlength{\tabcolsep}{1.8mm}
    \scalebox{0.9}{\begin{tabular}{c|c|c c c c c c } \toprule 
         $\balpha$&  $\bbeta$ &  R-18 &  R-50 & CvNt-B & CvNt-L & Swin-B & Swin-L\\ \midrule
         \ding{55} & \ding{55} & 64.0  & 71.6 & 81.0 & 81.4 & 79.8 & 82.4 \\ 
         \ding{55} & \ding{51} & 24.8 & 30.4 & 50.0 & 51.6 & 48.0 & 54.2\\  
         \ding{51} & \ding{55} & \underline{19.0} & \underline{25.4} & \underline{33.4} & \underline{35.4} & \underline{32.6} & \underline{36.0} \\  
        \ding{51}  & \ding{51} & \textbf{2.0} & \textbf{2.0} & \textbf{11.2} & \textbf{13.4} & \textbf{10.0} & \textbf{8.8} \\ \midrule
        \multicolumn{2}{c|}{Clean} & 69.6 & 77.4 & 84.2 & 84.0 & 81.6 & 82.4 \\\bottomrule
    \end{tabular}}
    \vspace{-7pt}
    \caption{\textbf{Classification accuracy (\%) under different adversarial diffusion guidance.} R-18/50 denotes the ResNet-18/50 networks and CvNt-B/L denotes the ConvNext-B/L networks.}
    \vspace{-10pt}
    \label{tab:ab}
\end{table}

\paragraph{Effect of perceptual constraint.} We use the LPIPS distance to bound the adversarial perturbations with a budget of $\gamma$. We show the effect of different $\gamma$ in~\Cref{tab:gamma} with ResNet-18 as the victim model. $\gamma=1$ corresponds to the unbounded case. As $\gamma$ increases, we have stronger attacks but lower FID~\cite{heusel2017gans}, which indicates lower image quality. Furthermore, a high $\gamma$ could lead to failures of semantic editing as shown in~\cref{fig:vis_gamma}. We set $\gamma=0.3$ since it achieves a nice tradeoff between image quality and attack performance. In addition, the average LPIPS distance between benign edited images and the original images is 0.29, which suggests that $\gamma=0.3$ is a reasonable LPIPS bound.

\begin{table}
    \centering
    \setlength{\tabcolsep}{1.8mm}
    \scalebox{0.9}{\begin{tabular}{c|cccccc}
    \toprule
        $\gamma$ & 0.1 & 0.2 & 0.3 & 0.5 &  0.7 &  1.0 \\ \midrule
        Accuracy (\%)& 15.6 & 3.6 & 2.0 & 0.2 & 0.2  & 0.2 \\
        FID &  34.71 & 44.03 & 56.72 & 73.35 & 80.83  & 82.28 \\ \bottomrule
    \end{tabular}}
    \vspace{-9pt}
    \caption{\textbf{Effect of $\gamma$.} }
    \label{tab:gamma}
    \vspace{-7pt}
\end{table}

\begin{figure}[t]
\centering
\small
\setlength{\tabcolsep}{0.2mm}
\scalebox{1.}{
\begin{tabular}[b]{ccccc}
    \begin{subfigure}{0.19\linewidth}
        \includegraphics[width=\textwidth]{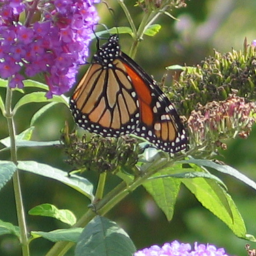}
        \caption*{Input}
    \end{subfigure} & 
    \begin{subfigure}{0.19\linewidth}
        \includegraphics[width=\textwidth]{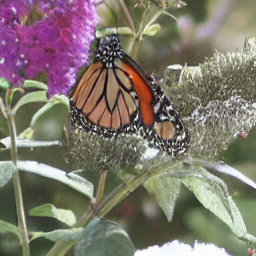}
        \caption*{$\gamma=0.1$}
    \end{subfigure} & 
    \begin{subfigure}{0.19\linewidth}
        \includegraphics[width=\textwidth]{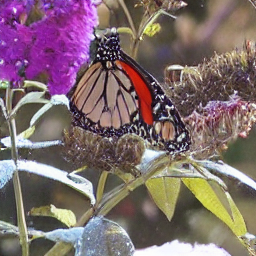}
        \caption*{$\gamma=0.3$}
    \end{subfigure} &
    \begin{subfigure}{0.19\linewidth}
        \includegraphics[width=\textwidth]{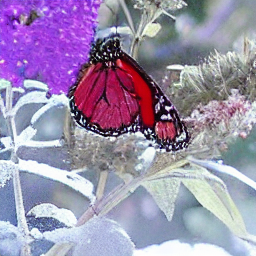}
        \caption*{$\gamma=0.5$}
    \end{subfigure} & 
    \begin{subfigure}{0.19\linewidth}
        \includegraphics[width=\textwidth]{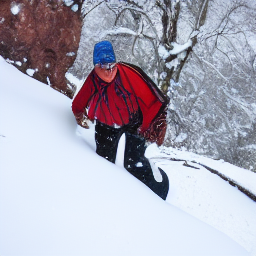}
        \caption*{$\gamma=1.0$}
    \end{subfigure}\\

\end{tabular}} 
\vspace{-10pt}
\caption{\textbf{Visualization} of adversarial samples with different $\gamma$.}
\vspace{-15pt}
\label{fig:vis_gamma}
\end{figure}

\section{Discussion and Conclusion}
In this work, we propose Instruct2Attack (I2A), a language-guided semantic attack that performs adversarial semantic editing on the input image based on the language instruction. I2A leverages a pretrained conditional latent diffusion model and adversarially guides the reverse diffusion process to search for an adversarial code in the latent space. The adversarial perturbation is bounded by the LPIPS distance to ensure the similarity between the input image and the adversarial image. We further explore to use GPT-4 to automatically generate diverse and image-specific instructions. Extensive experiments demonstrate the superiority of I2A under both white-box and black-box settings. 

The perturbations generated by I2A are natural, semantically meaningful, and interpretable. I2A reveals the vulnerabilities of DNNs under common natural semantic modifications. The success of I2A suggests the ubiquity of adversarial examples on the image manifold. Since on-manifold adversarial examples are generalization errors~\cite{stutz2019disentangling}, we believe the failures caused by I2A reflect the biases in datasets where certain semantic combinations are less common, such as nighttime and snowy scenes. I2A can serve as a flexible tool for evaluating model robustness under various semantic changes and diagnosing model failure modes. It can also be used as a data augmentation technique to generate synthetic training samples to enhance model robustness. 

\paragraph{\textbf{Acknowledgment}} This work was partially supported by the DARPA GARD Program HR001119S0026-GARD-FP-052. Jiang Liu acknowledges support through a fellowship from JHU + Amazon Initiative for Interactive AI (AI2AI).

{
    \small
    \bibliographystyle{ieeenat_fullname}
    \bibliography{main}
}

\appendix
\clearpage
\setcounter{page}{1}
\maketitlesupplementary
\renewcommand{\thefigure}{\Alph{figure}}
\renewcommand{\thetable}{\Alph{table}}
\setcounter{equation}{7}
\setcounter{table}{0}
\setcounter{figure}{0}
\section{Additional Implementation Details}
\subsection{Experiment Settings}
All experiments are conducted using the \texttt{PyTorch} framework. We fix the random seed of every experiment to avoid the randomness of diffusion sampling. 
\subsection{Projection Algorithm}
At the end of the attack generation, we ensure the perceptual constraint is satisfied by projecting $\vect{x}_{\text{adv}}$ back to the feasible set through dynamically adjusting $s_I$ and $s_T$ in~\cref{eq:cfg_adv} which control the trade-off between how strongly the edited image corresponds with the input image and the edit instruction. A higher $s_I$ produces an image that is more faithful to the input image, while a higher $s_T$ results in a stronger edit applied to the input. In this work, we use the default setting $s_I=1.5$ and $s_T=7.5$ suggested by~\cite{brooks2023instructpix2pix}. To project the adversarial image $\vect{x}_\text{adv}$ back to the LPIPS bound of $\gamma$, we first set $s_T=0$ and gradually increase the value of $s_I$ to find a value of $s_I^*$ such that the resulting image satisfies the LPIPS bound. Then we aim to increase the value of $s_T$ from 0 to find a $s_T^*$ that is as large as possible using the bisection method such that the final projected image satisfies the perceptual constraint. To search for $s_I^*$, we set the maximum value ${s_I}_\text{max}$ to 10, and the search step size $\eta_{s_I}$ to 0.2. To calculate $s_T^*$, we use $n=10$ iterations of the bisection method with the maximum value of $s_T$ set to ${s_T}_\text{max}=20$. The projection algorithm is summarized in~\Cref{alg:proj}. 

\begin{algorithm}[h]
\caption{Perceptual Projection}\label{alg:proj}
\begin{algorithmic}[1]
\State {\bfseries Input:} $\vect{x}$, $\vect{T}$, $\vect{z}_T$, $\gamma$, $T$, $\epsilon_\theta$, $\gE$, $\gD$, $\gC$,  optimized $\balpha$ and $\bbeta$, initial $s_I$, ${s_I}_\text{max}$, ${s_T}_\text{max}$, $\eta_{s_I}$, $n$.
\State {\bfseries Output:} Projected adversarial image $\Tilde{\vect{x}}_{\text{adv}}$ 
\LineComment{Image and text encoding}
\State $\vect{c}_I=\gE(\vect{x}),\ \ \vect{c}_L=\gC(\vect{L})$
\LineComment{Search for $s_I^*$} 
\State $s_T \gets 0$
\State {\bfseries while} $s_I \leq {s_I}_\text{max}$
\State $\ \ \ \ \Tilde{\vect{z}} =\text{LDM}(\vect{z}_T; \vect{c}_{I}, \vect{c}_{L}, \balpha, \bbeta, s_I, s_T$) \Comment{Sample $\Tilde{\vect{z}}$}
\State $\ \ \ \ \Tilde{\vect{x}}=\gD(\Tilde{\vect{z}})$ \Comment{Generate $\Tilde{\vect{x}}$}

\State \ \ \ \ {\bfseries if} $d(\Tilde{\vect{x}}, \vect{x})\leq \gamma$
\State \ \ \ \ \ \ \ \ $s_I^*=s_I$
\State \ \ \ \ \ \ \ \ {\bfseries break}
\State \ \ \ \ {\bfseries else}
\State \ \ \ \ \ \ \ \ $s_I = s_I + \eta_{s_I}$
\State \ \ \ \ {\bfseries end if} 
\State {\bfseries end while}

\LineComment{Calculate $s_T^*$} 
\State ${s_T}_\text{min} \gets 0$
\State {\bfseries for} $i=0$ \textbf{to} $n$
\State \ \ \ \  $s_T = ({s_T}_\text{min} + {s_T}_\text{max})/2$
\State $\ \ \ \ \Tilde{\vect{z}} =\text{LDM}(\vect{z}_T; \vect{c}_{I}, \vect{c}_{L}, \balpha, \bbeta, s_I^*, s_T$) \Comment{Sample $\Tilde{\vect{z}}$}
\State $\ \ \ \ \Tilde{\vect{x}}=\gD(\Tilde{\vect{z}})$ \Comment{Generate $\Tilde{\vect{x}}$}

\State \ \ \ \ {\bfseries if} $d(\Tilde{\vect{x}}, \vect{x}) > \gamma$
\State \ \ \ \ \ \ \ \ ${s_T}_\text{max}=s_T$
\State \ \ \ \ {\bfseries else}
\State \ \ \ \ \ \ \ \ ${s_T}_\text{min}=s_T$, $\Tilde{\vect{x}}_\text{adv} = \Tilde{\vect{x}}$
\State \ \ \ \ {\bfseries end if} 
\State {\bfseries end for}

\end{algorithmic}
\end{algorithm}

\subsection{Baseline Attacks}
The implementation details for the baseline attacks are as follows:
\begin{itemize}
    \item \textbf{FGSM}~\cite{szegedy2013intriguing}  We use $L_{\infty}$ norm with a bound of $4/255$.
    \item \textbf{PGD}~\cite{madry2018towards}:  We use $L_{\infty}$ norm with a bound of $4/255$ with 100 iterations and step size $1/255$.
    \item \textbf{MIM}~\cite{dong2018boosting}: We use $L_{\infty}$ norm with a bound of $4/255$ with 100 iterations, step size $1/255$, and decay factor $1.0$.
    \item \textbf{AutoAttack}~\cite{croce2020reliable}: We use $L_{\infty}$ norm with a bound of $4/255$ and use the \texttt{standard} version of AutoAttack\footnote{https://github.com/fra31/auto-attack}.
    \item \textbf{Fog}~\cite{kang2019testing}: We set the number of iterations to 200, $\epsilon$ to $128$, and step size to 0.002236.
    \item \textbf{Snow}~\cite{kang2019testing}: We set the number of iterations to 200, $\epsilon$ to $0.0625$, and step size to 0.002236.
    \item \textbf{Gabor}~\cite{kang2019testing}: We set the number of iterations to 200, $\epsilon$ to $12.5$, and step size to 0.002236.
    \item \textbf{StAdv}~\cite{xiao2018spatially}: We set the number of iterations to 200 with a bound of $0.005$.
    \item \textbf{PerC-AL}~\cite{zhao2020towards}: We follow the default settings in the official github repository\footnote{https://github.com/ZhengyuZhao/PerC-Adversarial}, and set the maximum number of iterations to 1000, $\alpha_l=1.0$, $\alpha_c=0.5$, and $\kappa=40$.
\end{itemize}
When applying EOT+BPDA attacks for adaptively attacking DiffPure in~\cref{sec:imgnet}, we set the maximum number of iterations to 50 for all attacks including I2A due to the high computational cost. 
\subsection{Victim Models for ImageNet}
The implementation details for the victim models of ImageNet are as follows:
\begin{itemize}
    \item \textbf{Undefended Models}: For ResNet-18/50, we use the pretrained models provided by \texttt{torchvision}. For Swin-B/L, we use the pretrained models provided in the official github repository\footnote{https://github.com/microsoft/Swin-Transformer}, with initial patch size of 4, window size of 7, and image size of 224. For ConvNext-B/L, we use the pretrained models provided in the official github repository\footnote{https://github.com/facebookresearch/ConvNeXt}.
    \item \textbf{Adversarially Trained Models}: For adversarially trained models, we use the model weights provided by Robustbench~\cite{croce2021robustbench}.
    \item \textbf{DiffPure Defended Models}: We apply the DiffPure~\cite{nie2022DiffPure} defense to the undefended models. We follow the implementations of the official github repository\footnote{https://github.com/NVlabs/DiffPure}. 
\end{itemize}
\subsection{Victim Models for Places365}
The implementation details for the victim models of Places365 are as follows:
\begin{itemize}
    \item \textbf{Undefended Models}: We use the pretrained ResNet-18/50 and DenseNet-161 models provided by the official repository of Places365\footnote{https://github.com/CSAILVision/places365}.
    \item \textbf{DiffPure Defended Models}: We apply the DiffPure~\cite{nie2022DiffPure} defense to the undefended models. 
\end{itemize}
\section{More Quantitative Results}
\subsection{Effect of $\lambda$}
We show the effect of the Lagrangian multiplier $\lambda$ of~\cref{eq:attack_lagrange} in~\Cref{tab:lambda}. A small $\lambda$ fails to regularize the output to satisfy the perceptual constraint, resulting in a high failure rate. We chose $\lambda=100$ since it achieves the lowest classification accuracy and failure rate.

\begin{table}[t]
    \centering
    \scalebox{1.0}{\begin{tabular}{c|cccccc}
    \toprule
        $\lambda$ & 10 & 50 & 100 & 150 &  200 &  500 \\ \midrule
        Accuracy (\%)& 24.6 & 2.8 & \textbf{1.8} & \textbf{1.8} & \textbf{1.8}  & 2.2 \\
        Failure (\%) &  28.8 & 1.4 & \textbf{0.0} & 0.2 & 0.4  & 0.2 \\ 
        \bottomrule
    \end{tabular}}
    \caption{\textbf{Effect of $\lambda$.} We report the top-1 accuracy and the failure rate where the LPIPS bound is not satisfied.}
    \label{tab:lambda}
\end{table}

\subsection{Comparison with Neural Perceptual Attacks}
By applying the perceptual constraint, I2A becomes a special case of the neural perceptual attacks (NPA)~\cite{laidlaw2021perceptual}, which aims to find an adversarial example ${\vect{x}_\text{adv}}\in \gX$ with a perceptual bound $\gamma$ such that ${\vect{x}_\text{adv}}$ is perceptually similar to $\vect{x}$ but causes the classifier $f$ to misclassify:
\begin{equation}
    f({\vect{x}_\text{adv}})\neq y,\quad\text{and}\quad d({\vect{x}_\text{adv}}, \vect{x})\leq \gamma,
    \label{eq:NPA}
\end{equation} 
where $d$ is the LPIPS distance, and $y$ is the ground truth label.~\cite{laidlaw2021perceptual} proposed the Lagrangian Perceptual Attack (LPA) to solve for ${\vect{x}_\text{adv}}$ by using a Lagrangian relaxation of the following constrained optimization problem:

\begin{equation}
    \max_{{\vect{x}_\text{adv}}} \mathcal{L}_m({\vect{x}_\text{adv}}, y),\  \text{s.t.},  d({\vect{x}_\text{adv}}, \vect{x}) \leq \gamma, 
    \label{eq:lpa}
\end{equation} 
where $\mathcal{L}_m$ is the margin loss from~\cite{carlini2017towards}. 

The main differences between LPA and I2A are as follows: 1) LPA is a \textit{noise-based } attack that generates $\vect{x}_\text{adv}$ by adding optimized adversarial noise to $\vect{x}$, while I2A is a \textit{semantic} attack that generates semantically meaningful perturbation and constraints $\vect{x}_\text{adv}$ to be \textit{on-manifold}; 2) The adversarial noise generated by LPA lacks interpretability and naturalness, while I2A generates \textit{interpretable} and \textit{natural} adversarial images; 2) I2A is a \textit{language-guided} attack that allows the user to control the adversarial perturbations through language instructions, while LPA does not provide such controllability and the resulting adversarial noise is purely driven by optimization. 

\begin{table}[t]
\centering
\scalebox{.9}{\begin{tabular}{c | l | c c c } 
\toprule
Model & Method & Clean &  LPA & I2A\\
\midrule
\multirow{3}{*}{\rotatebox{0}{ResNet-18}} & Undefended &69.68        & {0.00} & {1.20} \\
 & Salman2020~\cite{salman2020adversarially} & 52.92        & {0.00} & {0.54} \\
  & DiffPure~\cite{nie2022DiffPure} & 64.70        & {61.98} & {42.47}\\
\midrule
\multirow{5}{*}{\rotatebox{0}{ResNet-50}} & Undefended & 76.52        & {0.00} & {2.11} \\
 & Salman2020~\cite{salman2020adversarially} & 64.02        & {0.00} & {1.42}   \\
 & Engstrom2019~\cite{engstrom2019learning}  & 62.56        & {0.02} & {2.52}  \\
 & FastAT~\cite{Wong2020Fast} & 55.62        & {0.00} & {1.29}  \\
    & DiffPure~\cite{nie2022DiffPure} & 70.54        & 68.56 & {46.87}\\
\midrule
\multirow{4}{*}{\rotatebox{0}{ConvNeXt-B}} & Undefended & 83.36        & 0.00 & 8.93  \\
 & ARES~\cite{liu2023comprehensive} & 76.02        & 0.18 & {4.12}  \\
 & ConvStem~\cite{singh2023revisiting} & 75.90        & 0.78 & {3.62} \\
    & DiffPure~\cite{nie2022DiffPure}& 78.10        & 75.94 & {53.34}  \\
\midrule
\multirow{4}{*}{\rotatebox{0}{ConvNeXt-L}} & Undefended & 83.62        & 0.04 & 9.32  \\
 & ARES~\cite{liu2023comprehensive} & 78.02        & 2.26 & {4.95}\\
 & ConvStem~\cite{singh2023revisiting} & 77.00     & 0.18 & {4.91}\\
    & DiffPure~\cite{nie2022DiffPure}& 78.68        & 76.40 & {52.71} \\
\midrule
\multirow{3}{*}{\rotatebox{0}{Swin-B}} & Undefended & 81.98        & 0.00 & 8.81  \\
 & ARES~\cite{liu2023comprehensive} & 76.16        & 0.82 & {3.47}\\
    & DiffPure~\cite{nie2022DiffPure} & 76.60        & 75.00 & {50.65}\\
\midrule
\multirow{3}{*}{\rotatebox{0}{Swin-L}} & Undefended & 85.06        & 0.00 & 8.40 \\
 & ARES~\cite{liu2023comprehensive} & 78.92        & 1.40 & {4.98}  \\
    & DiffPure~\cite{nie2022DiffPure} & 81.40         & 79.86 & {55.18}\\
\midrule
\multicolumn{2}{c|}{Average Accuracy} & 73.97 & \second{20.16} & \best{16.90}\\ \bottomrule
\end{tabular}}
\caption{\textbf{Comparison with LPA under white-box attacks.} We report top-1 accuracy (\%) on ImageNet under white-box attacks.}
\label{tab:wb-lpa}
\end{table}

We provide quantitative comparisons between I2A and LPA under white-box attacks in~\Cref{tab:wb-lpa}. We set the perceptual bound to 0.3 for LPA and use the default settings of the official github repository\footnote{https://github.com/cassidylaidlaw/perceptual-advex}. LPA demonstrates strong attack ability on both undefended and adversarially trained models, resulting in nearly zero classification accuracy. However, since LPA is a noise-based attack, the adversarial noise can be easily removed by DiffPure, resulting in much lower attack performance than I2A when the DiffPure defense is applied. We also apply the EOT+BPDA techniques for adaptively attacking DiffPure and the results are summarized in~\Cref{tab:bpda-lpa}. I2A achieves significantly better performance than LPA under the DiffPure defense, especially under the adaptive attack setting. In addition, I2A achieves much better transferability than LPA under black-box attacks, as shown in~\cref{fig:blackbox-ipa}. For example, I2A achieves an average of 39.90\% classification accuracy for adversarially trained models, which is 14.91 points lower than LPA.

\begin{table}
\centering
\setlength{\tabcolsep}{2.0mm}
\scalebox{0.9}{\begin{tabular}{c|c| c c c} 
\toprule
Model & Adaptive & Clean & LPA & \textbf{I2A} \\
\midrule
\multirow{2}{*}{ResNet-18} & \ding{55} & 70.0 & 67.5 & \textbf{43.5} \\
 & \ding{51} & 70.0 & 34.0 & \textbf{7.0 }\\
 \midrule
\multirow{2}{*}{ResNet-50} & \ding{55} & 75.0 & 74.5 & \textbf{52.5} \\
 & \ding{51} &  75.0 & 46.0 &  \textbf{9.0}\\
 \midrule
\multirow{2}{*}{Swin-B} & \ding{55} & 80.0 & 75.0 & \textbf{52.0} \\
 & \ding{51} & 80.0 & 62.5 &  \textbf{20.5}\\
\bottomrule
\end{tabular}}
\caption{\textbf{DiffPure defense under adaptive and non-adaptive attacks}~\cite{nie2022DiffPure}. We report top-1 (\%) on ImageNet. }
\label{tab:bpda-lpa}
\end{table}

\begin{figure}[t]
\centering
\hspace*{-0.6cm}  
\small
\setlength{\tabcolsep}{0mm}
\scalebox{0.95}{
\begin{tabular}[b]{ cc}
    \begin{subfigure}{0.25\textwidth}
        \includegraphics[width=\textwidth]{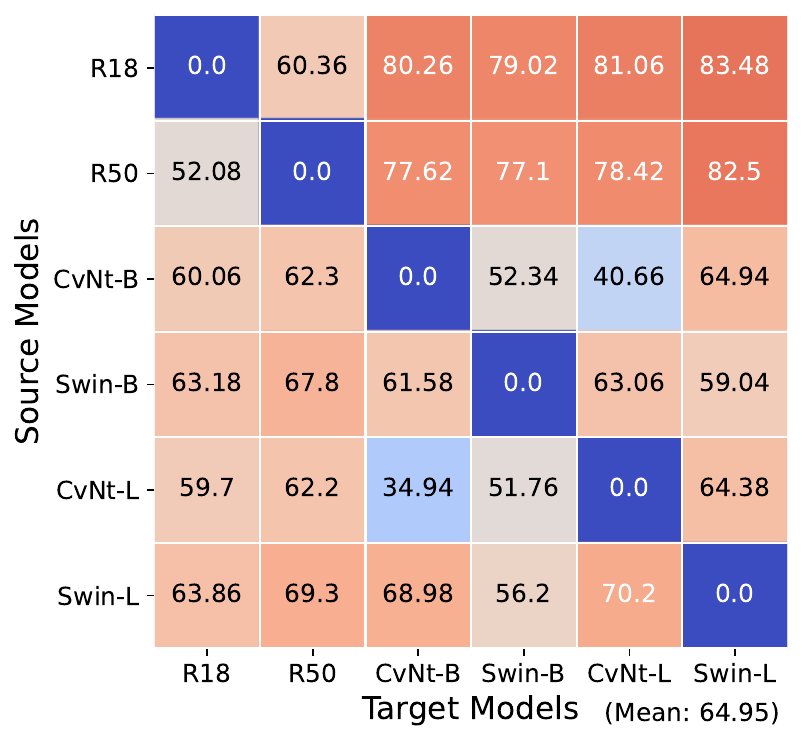}
    \end{subfigure} & 
    \begin{subfigure}{0.293\textwidth}
        \includegraphics[width=\textwidth]{figs/imgs/I2A_no_at.pdf}
    \end{subfigure}  \\
    \begin{subfigure}{0.25\textwidth}
        \includegraphics[width=\textwidth]{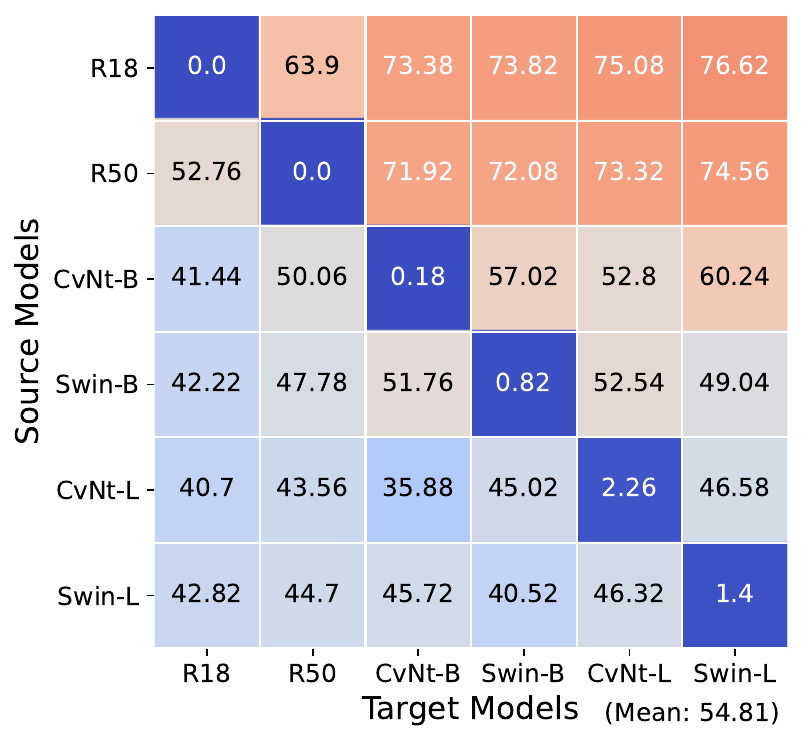}
        \caption{LPA~\cite{laidlaw2021perceptual}}
    \end{subfigure} & 
    \begin{subfigure}{0.293\textwidth}
        \includegraphics[width=\textwidth]{figs/imgs/I2A_at.pdf}
        \caption{\textbf{I2A (Ours)}}
    \end{subfigure}  \\
\end{tabular}}

\caption{\textbf{Black-box attacks.} We report top-1 (\%) on ImageNet. \textit{Top}: transferability between undefended models. \textit{Bottom}: transferability between adversarially trained models. Average accuracy of black-box attacks is marked on the bottom right of each figure.}
\vspace{-7pt}
\label{fig:blackbox-ipa}
\end{figure}

\subsection{Performance of Each Prompt}
We report the detailed performance of each prompt used by I2A in~\Cref{tab:wb-prompt}. We use four manually written prompts and one GPT-4 generated prompt. The five prompts achieve similar performance, which suggests the performance I2A is not sensitive to the input prompt. The prompt \texttt{"make it a vintage photo"} achieves slightly better performance compared to the others in terms of average accuracy.  
\begin{table*}[t]
\centering
\scalebox{1}{\begin{tabular}{c | l | c c c c c c | c} 
\toprule
Model & Method & Clean &  snow & night & painting & vintage & GPT-4 & Mean\\
\midrule
\multirow{3}{*}{\rotatebox{0}{ResNet-18}} & Undefended &69.68        & 0.82&0.94&\best{0.38}&\second{0.60}&3.26   & 1.20 \\
 & Salman2020~\cite{salman2020adversarially} & 52.92        &0.42&\second{0.28}&0.56&\best{0.22}&1.20 & {0.54} \\
  & DiffPure~\cite{nie2022DiffPure} & 64.70        & 41.84&43.80&45.08&\best{39.94}&\second{41.70} & {42.47}\\
\midrule
\multirow{5}{*}{\rotatebox{0}{ResNet-50}} & Undefended & 76.52        & 1.82&1.86&\best{0.82}&\second{0.96}&5.08 & 2.11 \\
 & Salman2020~\cite{salman2020adversarially} & 64.02        & 1.46&\best{0.90}&1.24&\second{0.98}&2.52& {1.42}   \\
 & Engstrom2019~\cite{engstrom2019learning}  & 62.56        & 2.60&\second{2.20}&2.54&\best{1.80}&3.44& {2.52}  \\
 & FastAT~\cite{Wong2020Fast} & 55.62        & 1.24&\second{0.92}&1.44&\best{0.90}&1.94 & {1.29}  \\
    & DiffPure~\cite{nie2022DiffPure} & 70.54        & 47.68&47.48&49.42&\best{44.00}&\second{45.78}& {46.87}\\
\midrule
\multirow{4}{*}{\rotatebox{0}{ConvNeXt-B}} & Undefended & 83.36        & 10.72&11.40&\best{6.02}&\second{6.42}&10.1 & 8.93  \\
 & ARES~\cite{liu2023comprehensive} & 76.02        & 5.00&4.38&\second{3.00}&\best{2.46}&5.74& {4.12}  \\
 & ConvStem~\cite{singh2023revisiting} & 75.90        & 4.62&4.06&\second{2.20}&\best{1.96}&5.26& {3.62} \\
    & DiffPure~\cite{nie2022DiffPure}& 78.10        & 54.82&53.86&55.36&\second{51.58}&\best{51.1}& {53.34}  \\
\midrule
\multirow{4}{*}{\rotatebox{0}{ConvNeXt-L}} & Undefended & 83.62        & 11.50&11.60&\second{6.76}&\best{6.14}&10.62& 9.32  \\
 & ARES~\cite{liu2023comprehensive} & 78.02        & 6.24&5.00&\second{3.32}&\best{3.30}&6.88& {4.95}\\
 & ConvStem~\cite{singh2023revisiting} & 77.00        & 6.16&5.38&\second{3.32}&\best{3.04}&6.66& {4.91}\\
    & DiffPure~\cite{nie2022DiffPure}& 78.68        &53.86&53.92&53.56&\second{52.00}&\best{50.22}& {52.71} \\
\midrule
\multirow{3}{*}{\rotatebox{0}{Swin-B}} & Undefended & 81.98        & 9.68&11.42&\best{5.52}&\second{6.12}&11.3& 8.81  \\
 & ARES~\cite{liu2023comprehensive} & 76.16        & 4.06&\second{3.18}&\best{2.06}&3.62&4.44& {3.47}\\
    & DiffPure~\cite{nie2022DiffPure} & 76.60        & 51.58&50.66&52.72&\second{49.26}&\best{49.04}& {50.65}\\
\midrule
\multirow{3}{*}{\rotatebox{0}{Swin-L}} & Undefended & 85.06        & 8.94&9.84&\second{6.34}&\best{6.04}&10.84& 8.40 \\
 & ARES~\cite{liu2023comprehensive} & 78.92        & 5.90&4.98&\best{3.56}&\second{3.62}&6.84 & {4.98}  \\
    & DiffPure~\cite{nie2022DiffPure} & 81.40         & 57.18&55.88&55.68&\second{54.12}&\best{53.02}& {55.18}\\
\midrule
\multicolumn{2}{c|}{Average Accuracy} & 73.97 & 17.64&17.45&\second{16.40}&\best{15.41}&17.59 & {16.90}\\ \bottomrule
\end{tabular}}
\caption{\textbf{Effect of Prompts.} We report top-1 accuracy (\%) on ImageNet under white-box attacks. We use four manually written prompts: \texttt{"make it in snow"} (snow), \texttt{"make it at night"} (night), \texttt{"make it a sketch painting"} (painting), and \texttt{"make it a vintage photo"} (vintage), as well as one prompt generated automatically by GPT-4. The best prompt of each row is in \best{bold} and the second best is \second{{underlined}}.}
\label{tab:wb-prompt}
\end{table*}

\section{More Visualization Results}
\begin{figure*}[t]
\centering
\small
\setlength{\tabcolsep}{1mm}
\scalebox{0.95}{
\begin{tabular}[b]{ cccccc}
    \begin{subfigure}{0.16\textwidth}
        \includegraphics[width=\textwidth]{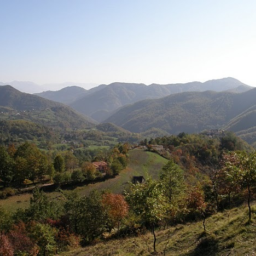}
        \caption*{Clean Image}
    \end{subfigure} & 
    \begin{subfigure}{0.16\textwidth}
        \includegraphics[width=\textwidth]{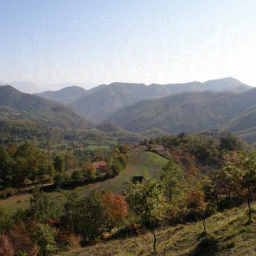}
        \caption*{PGD~\cite{madry2018towards}}
    \end{subfigure} & 
    \begin{subfigure}{0.16\textwidth}
        \includegraphics[width=\textwidth]{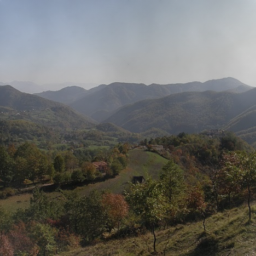}
        \caption*{Fog~\cite{kang2019testing}}
    \end{subfigure} & 
    \begin{subfigure}{0.16\textwidth}
        \includegraphics[width=\textwidth]{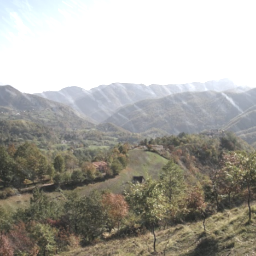}
        \caption*{Snow~\cite{kang2019testing}}
    \end{subfigure} & 
    \begin{subfigure}{0.16\textwidth}
        \includegraphics[width=\textwidth]{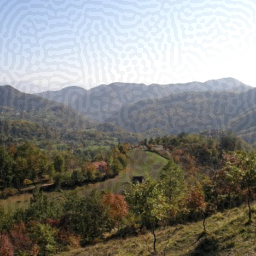}
        \caption*{Gabor~\cite{kang2019testing}}
    \end{subfigure} & 
    \begin{subfigure}{0.16\textwidth}
        \includegraphics[width=\textwidth]{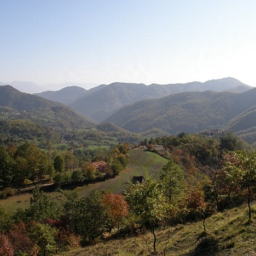}
        \caption*{PerC-AL~\cite{zhang2018perceptual}}
    \end{subfigure} \\
    \begin{subfigure}[t]{0.16\textwidth}
        \includegraphics[width=\textwidth]{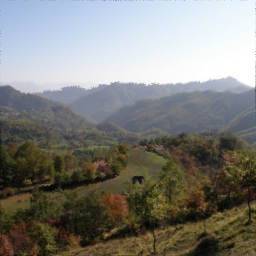}
        \caption*{StAdv~\cite{xiao2018spatially}}
    \end{subfigure} & 
    \begin{subfigure}[t]{0.16\textwidth}
        \includegraphics[width=\textwidth]{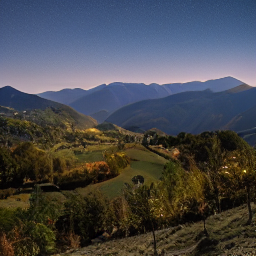}
        \caption*{\textbf{I2A:} \textit{``make it at night"}}
    \end{subfigure} & 
    \begin{subfigure}[t]{0.16\textwidth}
        \includegraphics[width=\textwidth]{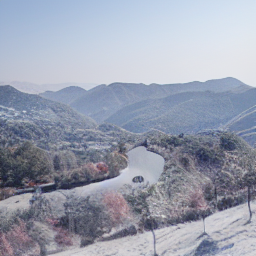}
        \caption*{\textbf{I2A:} \textit{``make it in snow"}}
    \end{subfigure} & 
    \begin{subfigure}[t]{0.16\textwidth}
        \includegraphics[width=\textwidth]{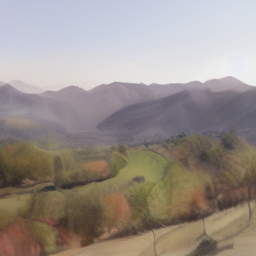}
        \caption*{\textbf{I2A:} \textit{``make it a sketch painting"}}
    \end{subfigure} & 
    \begin{subfigure}[t]{0.16\textwidth}
        \includegraphics[width=\textwidth]{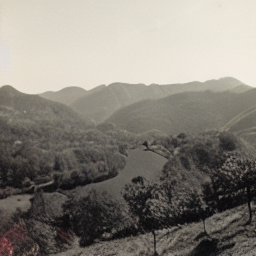}
        \caption*{\textbf{I2A:} \textit{``make it a vintage photo"}}
    \end{subfigure} & 
    \begin{subfigure}[t]{0.16\textwidth}
        \includegraphics[width=\textwidth]{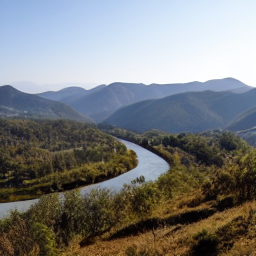}
        \caption*{\textbf{I2A:} \textit{``Add a river in the valley"} (GPT-4)}
     \end{subfigure}  \\ \hline 
    \vspace{-5pt}\\
    \begin{subfigure}[b]{0.16\textwidth}
        \includegraphics[width=\textwidth]{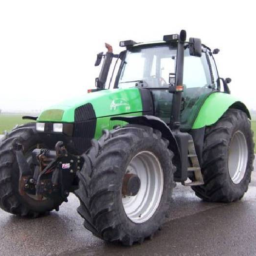}
        \caption*{Clean Image}
    \end{subfigure} & 
    \begin{subfigure}[b]{0.16\textwidth}
        \includegraphics[width=\textwidth]{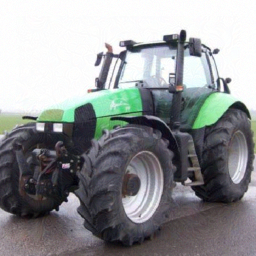}
        \caption*{PGD~\cite{madry2018towards}}
    \end{subfigure} & 
    \begin{subfigure}[b]{0.16\textwidth}
        \includegraphics[width=\textwidth]{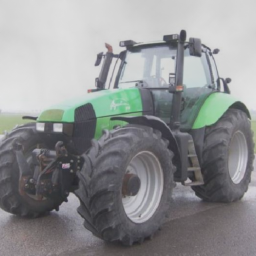}
        \caption*{Fog~\cite{kang2019testing}}
    \end{subfigure} & 
    \begin{subfigure}[b]{0.16\textwidth}
        \includegraphics[width=\textwidth]{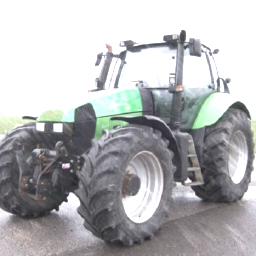}
        \caption*{Snow~\cite{kang2019testing}}
    \end{subfigure} & 
    \begin{subfigure}[b]{0.16\textwidth}
        \includegraphics[width=\textwidth]{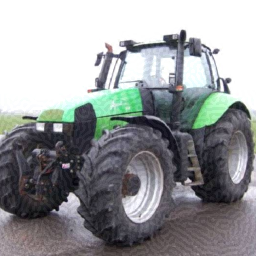}
        \caption*{Gabor~\cite{kang2019testing}}
    \end{subfigure} & 
    \begin{subfigure}[b]{0.16\textwidth}
        \includegraphics[width=\textwidth]{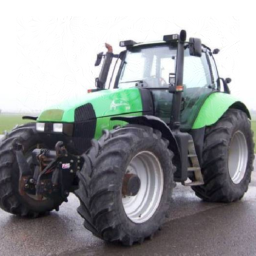}
        \caption*{PerC-AL~\cite{zhao2020towards}}
    \end{subfigure} \\
    \begin{subfigure}[t]{0.16\textwidth}
        \includegraphics[width=\textwidth]{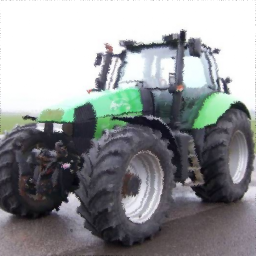}
        \caption*{StAdv~\cite{xiao2018spatially}}
    \end{subfigure} & 
    \begin{subfigure}[t]{0.16\textwidth}
        \includegraphics[width=\textwidth]{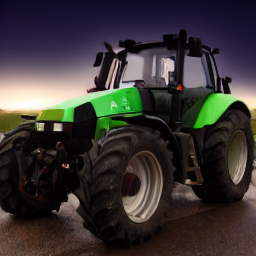}
        \caption*{\textbf{I2A:} \textit{``make it at night"}}
    \end{subfigure} & 
    \begin{subfigure}[t]{0.16\textwidth}
        \includegraphics[width=\textwidth]{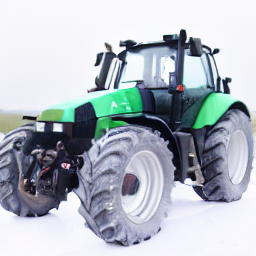}
        \caption*{\textbf{I2A:} \textit{``make it in snow"}}
    \end{subfigure} & 
    \begin{subfigure}[t]{0.16\textwidth}
        \includegraphics[width=\textwidth]{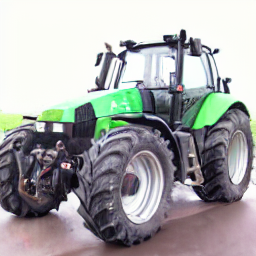}
        \caption*{\textbf{I2A:} \textit{``make it a sketch painting"}}
    \end{subfigure} & 
    \begin{subfigure}[t]{0.16\textwidth}
        \includegraphics[width=\textwidth]{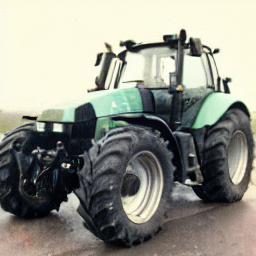}
        \caption*{\textbf{I2A:} \textit{``make it a vintage photo"}}
    \end{subfigure} & 
    \begin{subfigure}[t]{0.16\textwidth}
        \includegraphics[width=\textwidth]{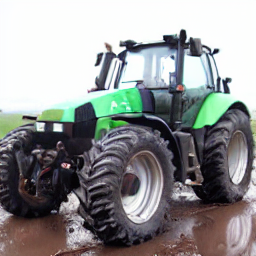}
        \caption*{\textbf{I2A:} \textit{``Add some mud on the tires"} (GPT-4)}
    \end{subfigure} \\ \hline 
    \vspace{-5pt}\\
    \begin{subfigure}[b]{0.16\textwidth}
        \includegraphics[width=\textwidth]{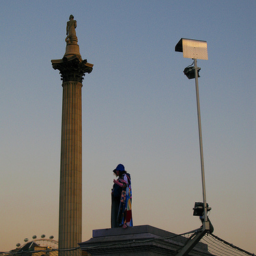}
        \caption*{Clean Image}
    \end{subfigure} & 
    \begin{subfigure}[b]{0.16\textwidth}
        \includegraphics[width=\textwidth]{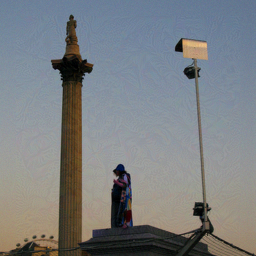}
        \caption*{PGD~\cite{madry2018towards}}
    \end{subfigure} & 
    \begin{subfigure}[b]{0.16\textwidth}
        \includegraphics[width=\textwidth]{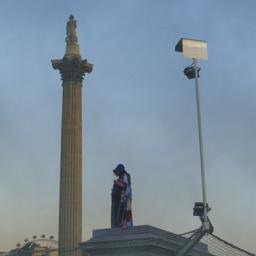}
        \caption*{Fog~\cite{kang2019testing}}
    \end{subfigure} & 
    \begin{subfigure}[b]{0.16\textwidth}
        \includegraphics[width=\textwidth]{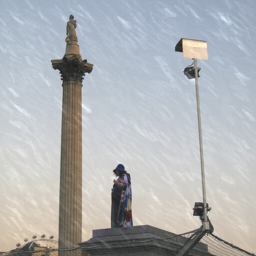}
        \caption*{Snow~\cite{kang2019testing}}
    \end{subfigure} & 
    \begin{subfigure}[b]{0.16\textwidth}
        \includegraphics[width=\textwidth]{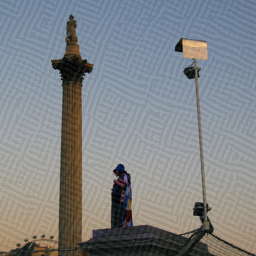}
        \caption*{Gabor~\cite{kang2019testing}}
    \end{subfigure} & 
    \begin{subfigure}[b]{0.16\textwidth}
        \includegraphics[width=\textwidth]{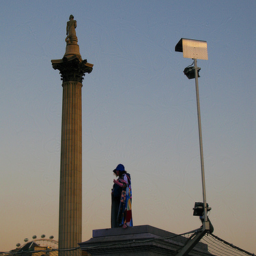}
        \caption*{PerC-AL~\cite{zhao2020towards}}
    \end{subfigure} \\
    \begin{subfigure}[t]{0.16\textwidth}
        \includegraphics[width=\textwidth]{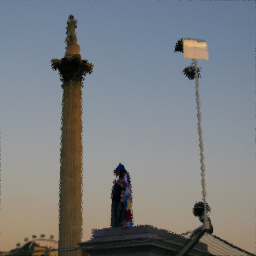}
        \caption*{StAdv~\cite{xiao2018spatially}}
    \end{subfigure} & 
    \begin{subfigure}[t]{0.16\textwidth}
        \includegraphics[width=\textwidth]{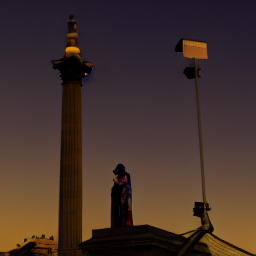}
        \caption*{\textbf{I2A:} \textit{``make it at night"}}
    \end{subfigure} & 
    \begin{subfigure}[t]{0.16\textwidth}
        \includegraphics[width=\textwidth]{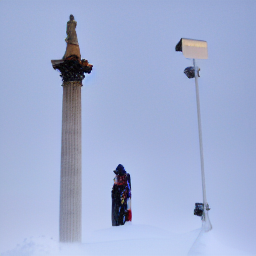}
        \caption*{\textbf{I2A:} \textit{``make it in snow"}}
    \end{subfigure} & 
    \begin{subfigure}[t]{0.16\textwidth}
        \includegraphics[width=\textwidth]{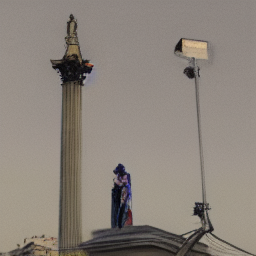}
        \caption*{\textbf{I2A:} \textit{``make it a sketch painting"}}
    \end{subfigure} & 
    \begin{subfigure}[t]{0.16\textwidth}
        \includegraphics[width=\textwidth]{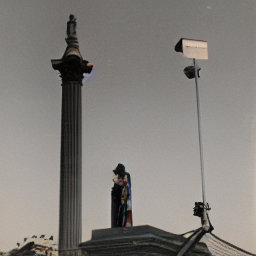}
        \caption*{\textbf{I2A:} \textit{``make it a vintage photo"}}
    \end{subfigure} & 
    \begin{subfigure}[t]{0.16\textwidth}
        \includegraphics[width=\textwidth]{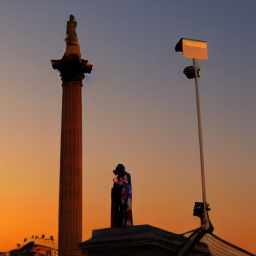}
        \caption*{\textbf{I2A:} \textit{``make it sunset"} (GPT-4)}
    \end{subfigure} \\
\end{tabular}} 
\caption{\textbf{Visualization of adversarial images on ImageNet.} I2A generates natural and diverse perturbations based on the text instructions.}
\label{fig:vis_supp-2}
\end{figure*}

\begin{figure*}[t]
\centering
\small
\setlength{\tabcolsep}{1mm}
\scalebox{0.95}{
\begin{tabular}[b]{ cccccc}
    \begin{subfigure}{0.16\textwidth}
        \includegraphics[width=\textwidth]{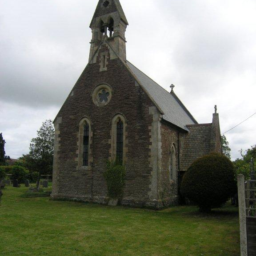}
        \caption*{Clean Image}
    \end{subfigure} & 
    \begin{subfigure}{0.16\textwidth}
        \includegraphics[width=\textwidth]{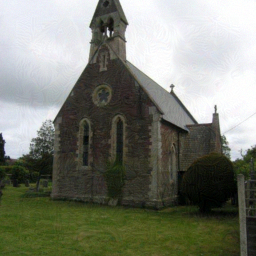}
        \caption*{PGD~\cite{madry2018towards}}
    \end{subfigure} & 
    \begin{subfigure}{0.16\textwidth}
        \includegraphics[width=\textwidth]{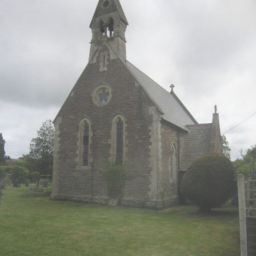}
        \caption*{Fog~\cite{kang2019testing}}
    \end{subfigure} & 
    \begin{subfigure}{0.16\textwidth}
        \includegraphics[width=\textwidth]{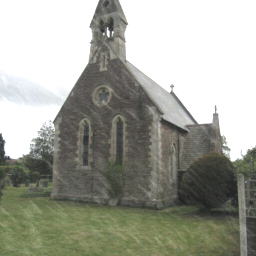}
        \caption*{Snow~\cite{kang2019testing}}
    \end{subfigure} & 
    \begin{subfigure}{0.16\textwidth}
        \includegraphics[width=\textwidth]{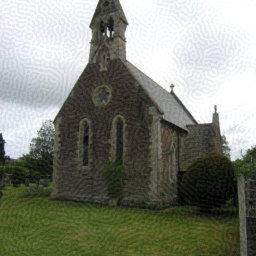}
        \caption*{Gabor~\cite{kang2019testing}}
    \end{subfigure} & 
    \begin{subfigure}{0.16\textwidth}
        \includegraphics[width=\textwidth]{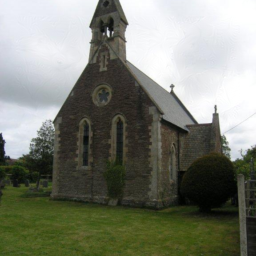}
        \caption*{PerC-AL~\cite{zhang2018perceptual}}
    \end{subfigure} \\
    \begin{subfigure}[t]{0.16\textwidth}
        \includegraphics[width=\textwidth]{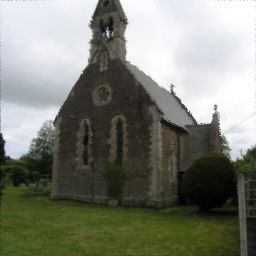}
        \caption*{StAdv~\cite{xiao2018spatially}}
    \end{subfigure} & 
    \begin{subfigure}[t]{0.16\textwidth}
        \includegraphics[width=\textwidth]{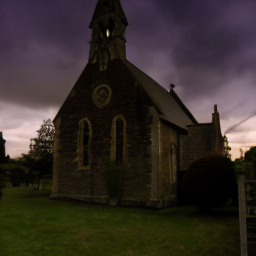}
        \caption*{\textbf{I2A:} \textit{``make it at night"}}
    \end{subfigure} & 
    \begin{subfigure}[t]{0.16\textwidth}
        \includegraphics[width=\textwidth]{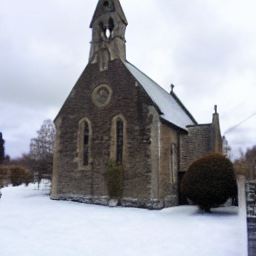}
        \caption*{\textbf{I2A:} \textit{``make it in snow"}}
    \end{subfigure} & 
    \begin{subfigure}[t]{0.16\textwidth}
        \includegraphics[width=\textwidth]{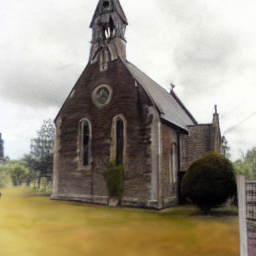}
        \caption*{\textbf{I2A:} \textit{``make it a sketch painting"}}
    \end{subfigure} & 
    \begin{subfigure}[t]{0.16\textwidth}
        \includegraphics[width=\textwidth]{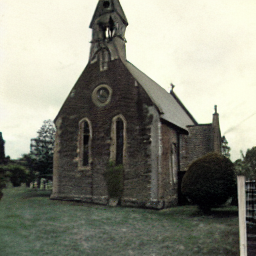}
        \caption*{\textbf{I2A:} \textit{``make it a vintage photo"}}
    \end{subfigure} & 
    \begin{subfigure}[t]{0.16\textwidth}
        \includegraphics[width=\textwidth]{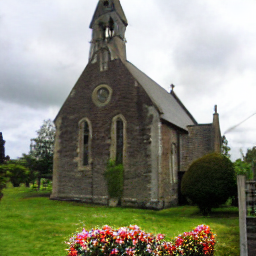}
        \caption*{\textbf{I2A:} \textit{``Add flowers around the church"} (GPT-4)}
     \end{subfigure}  \\ \hline 
    \vspace{-5pt}\\
    \begin{subfigure}[b]{0.16\textwidth}
        \includegraphics[width=\textwidth]{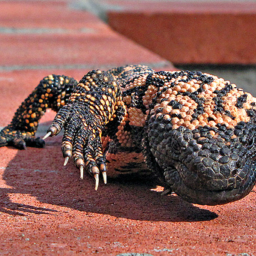}
        \caption*{Clean Image}
    \end{subfigure} & 
    \begin{subfigure}[b]{0.16\textwidth}
        \includegraphics[width=\textwidth]{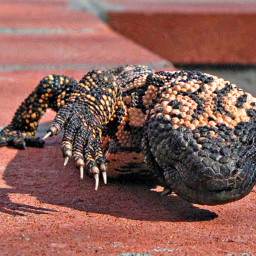}
        \caption*{PGD~\cite{madry2018towards}}
    \end{subfigure} & 
    \begin{subfigure}[b]{0.16\textwidth}
        \includegraphics[width=\textwidth]{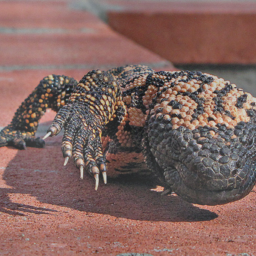}
        \caption*{Fog~\cite{kang2019testing}}
    \end{subfigure} & 
    \begin{subfigure}[b]{0.16\textwidth}
        \includegraphics[width=\textwidth]{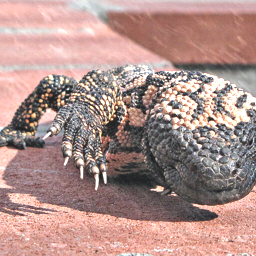}
        \caption*{Snow~\cite{kang2019testing}}
    \end{subfigure} & 
    \begin{subfigure}[b]{0.16\textwidth}
        \includegraphics[width=\textwidth]{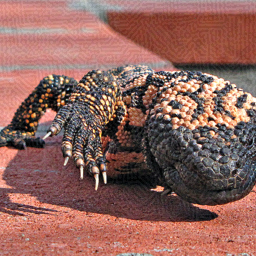}
        \caption*{Gabor~\cite{kang2019testing}}
    \end{subfigure} & 
    \begin{subfigure}[b]{0.16\textwidth}
        \includegraphics[width=\textwidth]{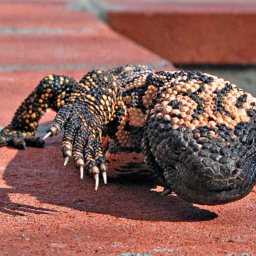}
        \caption*{PerC-AL~\cite{zhao2020towards}}
    \end{subfigure} \\
    \begin{subfigure}[t]{0.16\textwidth}
        \includegraphics[width=\textwidth]{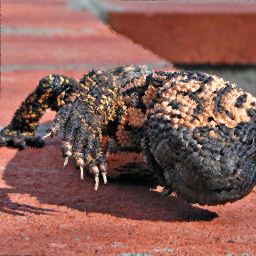}
        \caption*{StAdv~\cite{xiao2018spatially}}
    \end{subfigure} & 
    \begin{subfigure}[t]{0.16\textwidth}
        \includegraphics[width=\textwidth]{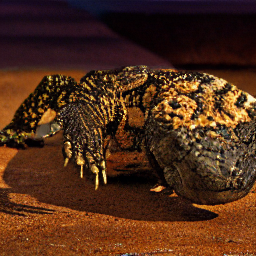}
        \caption*{\textbf{I2A:} \textit{``make it at night"}}
    \end{subfigure} & 
    \begin{subfigure}[t]{0.16\textwidth}
        \includegraphics[width=\textwidth]{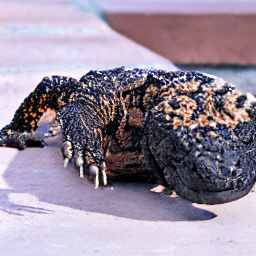}
        \caption*{\textbf{I2A:} \textit{``make it in snow"}}
    \end{subfigure} & 
    \begin{subfigure}[t]{0.16\textwidth}
        \includegraphics[width=\textwidth]{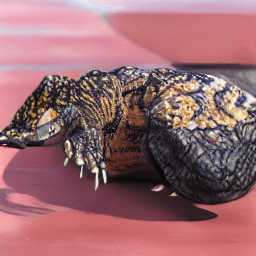}
        \caption*{\textbf{I2A:} \textit{``make it a sketch painting"}}
    \end{subfigure} & 
    \begin{subfigure}[t]{0.16\textwidth}
        \includegraphics[width=\textwidth]{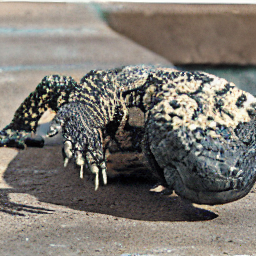}
        \caption*{\textbf{I2A:} \textit{``make it a vintage photo"}}
    \end{subfigure} & 
    \begin{subfigure}[t]{0.16\textwidth}
        \includegraphics[width=\textwidth]{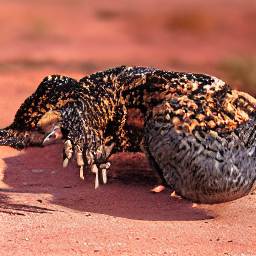}
        \caption*{\textbf{I2A:} \textit{``In a desert location"} (GPT-4)}
    \end{subfigure} \\ \hline 
    \vspace{-5pt}\\
    \begin{subfigure}[b]{0.16\textwidth}
        \includegraphics[width=\textwidth]{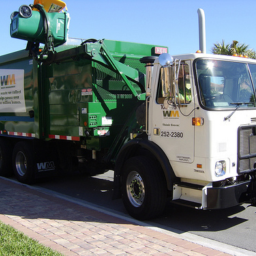}
        \caption*{Clean Image}
    \end{subfigure} & 
    \begin{subfigure}[b]{0.16\textwidth}
        \includegraphics[width=\textwidth]{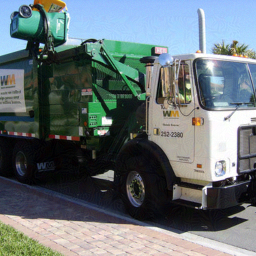}
        \caption*{PGD~\cite{madry2018towards}}
    \end{subfigure} & 
    \begin{subfigure}[b]{0.16\textwidth}
        \includegraphics[width=\textwidth]{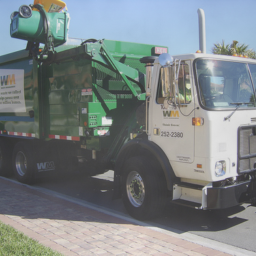}
        \caption*{Fog~\cite{kang2019testing}}
    \end{subfigure} & 
    \begin{subfigure}[b]{0.16\textwidth}
        \includegraphics[width=\textwidth]{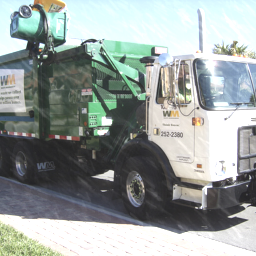}
        \caption*{Snow~\cite{kang2019testing}}
    \end{subfigure} & 
    \begin{subfigure}[b]{0.16\textwidth}
        \includegraphics[width=\textwidth]{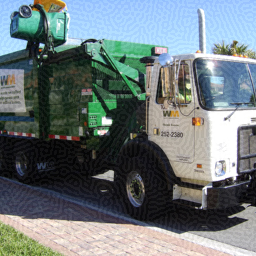}
        \caption*{Gabor~\cite{kang2019testing}}
    \end{subfigure} & 
    \begin{subfigure}[b]{0.16\textwidth}
        \includegraphics[width=\textwidth]{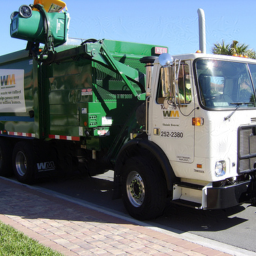}
        \caption*{PerC-AL~\cite{zhao2020towards}}
    \end{subfigure} \\
    \begin{subfigure}[t]{0.16\textwidth}
        \includegraphics[width=\textwidth]{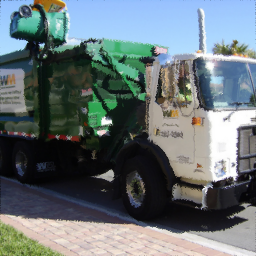}
        \caption*{StAdv~\cite{xiao2018spatially}}
    \end{subfigure} & 
    \begin{subfigure}[t]{0.16\textwidth}
        \includegraphics[width=\textwidth]{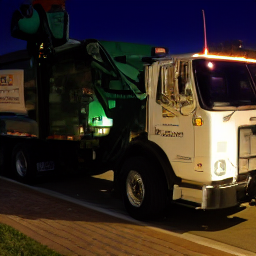}
        \caption*{\textbf{I2A:} \textit{``make it at night"}}
    \end{subfigure} & 
    \begin{subfigure}[t]{0.16\textwidth}
        \includegraphics[width=\textwidth]{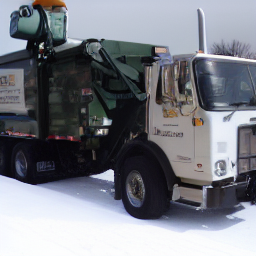}
        \caption*{\textbf{I2A:} \textit{``make it in snow"}}
    \end{subfigure} & 
    \begin{subfigure}[t]{0.16\textwidth}
        \includegraphics[width=\textwidth]{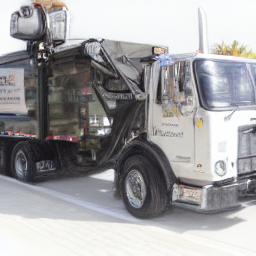}
        \caption*{\textbf{I2A:} \textit{``make it a sketch painting"}}
    \end{subfigure} & 
    \begin{subfigure}[t]{0.16\textwidth}
        \includegraphics[width=\textwidth]{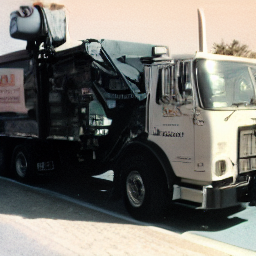}
        \caption*{\textbf{I2A:} \textit{``make it a vintage photo"}}
    \end{subfigure} & 
    \begin{subfigure}[t]{0.16\textwidth}
        \includegraphics[width=\textwidth]{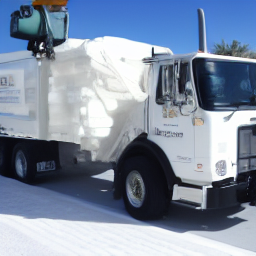}
        \caption*{\textbf{I2A:} \textit{``make it fully white"} (GPT-4)}
    \end{subfigure} \\
\end{tabular}} 
\caption{\textbf{Visualization of adversarial images on ImageNet.} I2A generates natural and diverse perturbations based on the text instructions.}
\label{fig:vis_supp}
\end{figure*}

\begin{figure*}[t]
\centering
\small
\setlength{\tabcolsep}{1mm}
\scalebox{0.95}{
\begin{tabular}[b]{ cccccc}
    \begin{subfigure}{0.16\textwidth}
        \includegraphics[width=\textwidth]{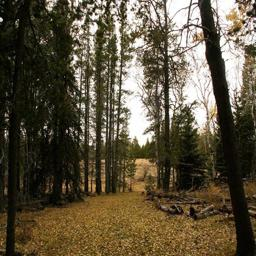}
        \caption*{Clean Image}
    \end{subfigure} & 
    \begin{subfigure}{0.16\textwidth}
        \includegraphics[width=\textwidth]{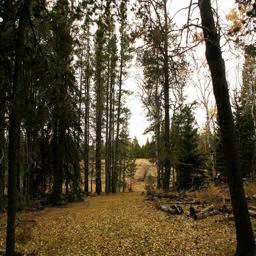}
        \caption*{PGD~\cite{madry2018towards}}
    \end{subfigure} & 
    \begin{subfigure}{0.16\textwidth}
        \includegraphics[width=\textwidth]{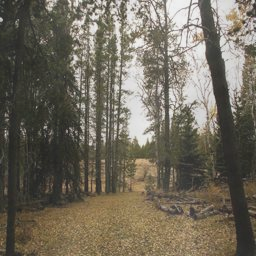}
        \caption*{Fog~\cite{kang2019testing}}
    \end{subfigure} & 
    \begin{subfigure}{0.16\textwidth}
        \includegraphics[width=\textwidth]{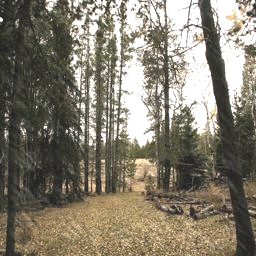}
        \caption*{Snow~\cite{kang2019testing}}
    \end{subfigure} & 
    \begin{subfigure}{0.16\textwidth}
        \includegraphics[width=\textwidth]{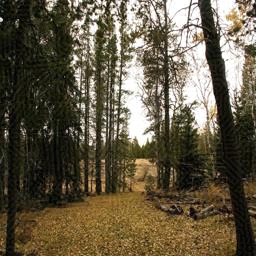}
        \caption*{Gabor~\cite{kang2019testing}}
    \end{subfigure} & 
    \begin{subfigure}{0.16\textwidth}
        \includegraphics[width=\textwidth]{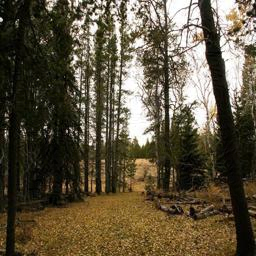}
        \caption*{PerC-AL~\cite{zhang2018perceptual}}
    \end{subfigure} \\
    \begin{subfigure}[t]{0.16\textwidth}
        \includegraphics[width=\textwidth]{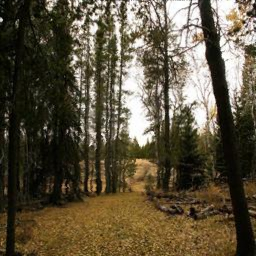}
        \caption*{StAdv~\cite{xiao2018spatially}}
    \end{subfigure} & 
    \begin{subfigure}[t]{0.16\textwidth}
        \includegraphics[width=\textwidth]{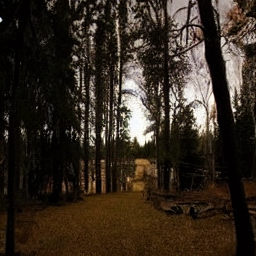}
        \caption*{\textbf{I2A:} \textit{``make it at night"}}
    \end{subfigure} & 
    \begin{subfigure}[t]{0.16\textwidth}
        \includegraphics[width=\textwidth]{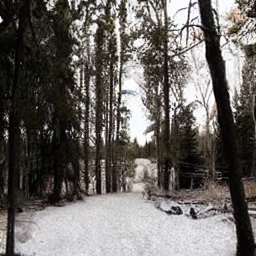}
        \caption*{\textbf{I2A:} \textit{``make it in snow"}}
    \end{subfigure} & 
    \begin{subfigure}[t]{0.16\textwidth}
        \includegraphics[width=\textwidth]{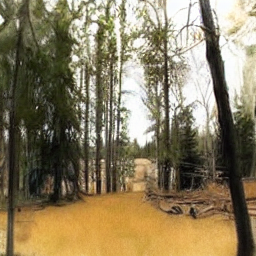}
        \caption*{\textbf{I2A:} \textit{``make it a sketch painting"}}
    \end{subfigure} & 
    \begin{subfigure}[t]{0.16\textwidth}
        \includegraphics[width=\textwidth]{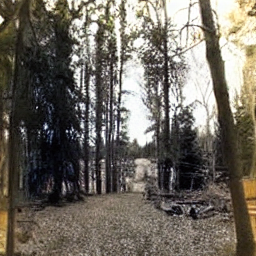}
        \caption*{\textbf{I2A:} \textit{``make it a vintage photo"}}
    \end{subfigure} & 
    \begin{subfigure}[t]{0.16\textwidth}
        \includegraphics[width=\textwidth]{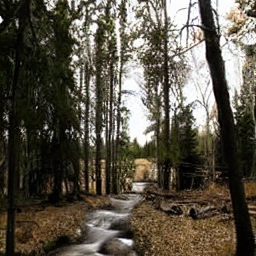}
        \caption*{\textbf{I2A:} \textit{``Add a stream"} (GPT-4)}
     \end{subfigure}  \\ \hline 
    \vspace{-5pt}\\
    \begin{subfigure}[b]{0.16\textwidth}
        \includegraphics[width=\textwidth]{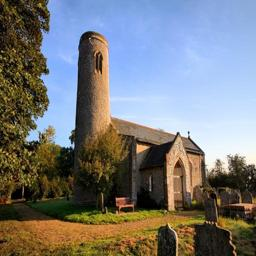}
        \caption*{Clean Image}
    \end{subfigure} & 
    \begin{subfigure}[b]{0.16\textwidth}
        \includegraphics[width=\textwidth]{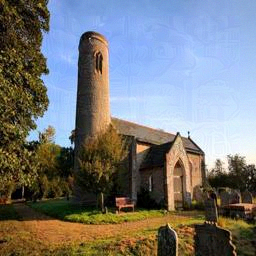}
        \caption*{PGD~\cite{madry2018towards}}
    \end{subfigure} & 
    \begin{subfigure}[b]{0.16\textwidth}
        \includegraphics[width=\textwidth]{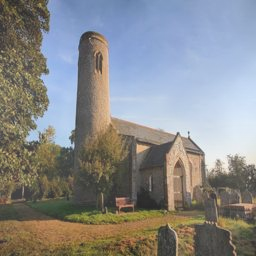}
        \caption*{Fog~\cite{kang2019testing}}
    \end{subfigure} & 
    \begin{subfigure}[b]{0.16\textwidth}
        \includegraphics[width=\textwidth]{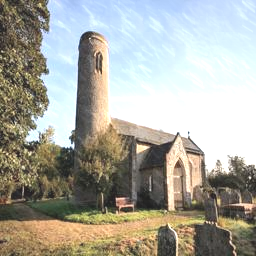}
        \caption*{Snow~\cite{kang2019testing}}
    \end{subfigure} & 
    \begin{subfigure}[b]{0.16\textwidth}
        \includegraphics[width=\textwidth]{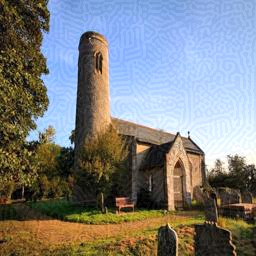}
        \caption*{Gabor~\cite{kang2019testing}}
    \end{subfigure} & 
    \begin{subfigure}[b]{0.16\textwidth}
        \includegraphics[width=\textwidth]{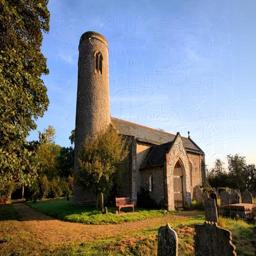}
        \caption*{PerC-AL~\cite{zhao2020towards}}
    \end{subfigure} \\
    \begin{subfigure}[t]{0.16\textwidth}
        \includegraphics[width=\textwidth]{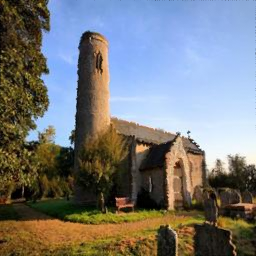}
        \caption*{StAdv~\cite{xiao2018spatially}}
    \end{subfigure} & 
    \begin{subfigure}[t]{0.16\textwidth}
        \includegraphics[width=\textwidth]{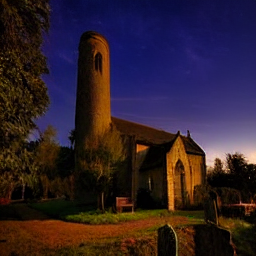}
        \caption*{\textbf{I2A:} \textit{``make it at night"}}
    \end{subfigure} & 
    \begin{subfigure}[t]{0.16\textwidth}
        \includegraphics[width=\textwidth]{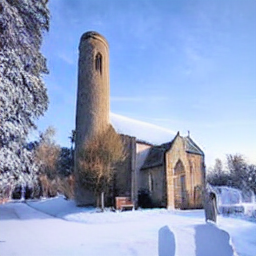}
        \caption*{\textbf{I2A:} \textit{``make it in snow"}}
    \end{subfigure} & 
    \begin{subfigure}[t]{0.16\textwidth}
        \includegraphics[width=\textwidth]{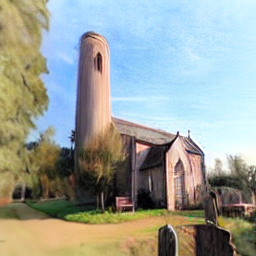}
        \caption*{\textbf{I2A:} \textit{``make it a sketch painting"}}
    \end{subfigure} & 
    \begin{subfigure}[t]{0.16\textwidth}
        \includegraphics[width=\textwidth]{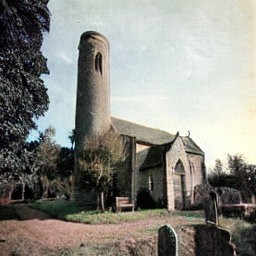}
        \caption*{\textbf{I2A:} \textit{``make it a vintage photo"}}
    \end{subfigure} & 
    \begin{subfigure}[t]{0.16\textwidth}
        \includegraphics[width=\textwidth]{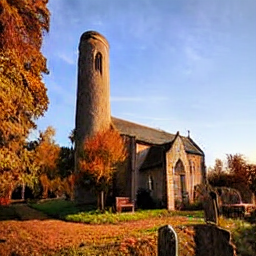}
        \caption*{\textbf{I2A:} \textit{``Make it Autumn time"} (GPT-4)}
    \end{subfigure} \\ \hline 
    \vspace{-5pt}\\
    \begin{subfigure}[b]{0.16\textwidth}
        \includegraphics[width=\textwidth]{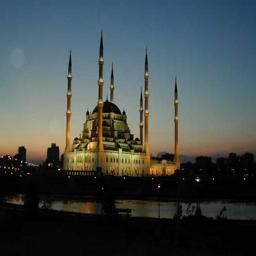}
        \caption*{Clean Image}
    \end{subfigure} & 
    \begin{subfigure}[b]{0.16\textwidth}
        \includegraphics[width=\textwidth]{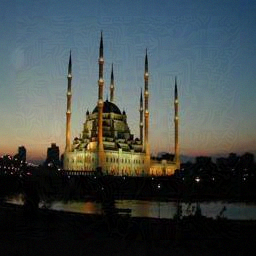}
        \caption*{PGD~\cite{madry2018towards}}
    \end{subfigure} & 
    \begin{subfigure}[b]{0.16\textwidth}
        \includegraphics[width=\textwidth]{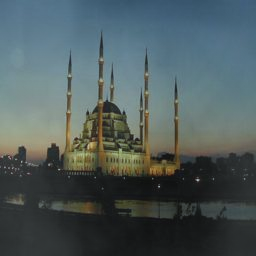}
        \caption*{Fog~\cite{kang2019testing}}
    \end{subfigure} & 
    \begin{subfigure}[b]{0.16\textwidth}
        \includegraphics[width=\textwidth]{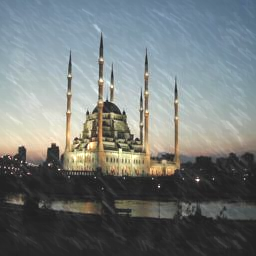}
        \caption*{Snow~\cite{kang2019testing}}
    \end{subfigure} & 
    \begin{subfigure}[b]{0.16\textwidth}
        \includegraphics[width=\textwidth]{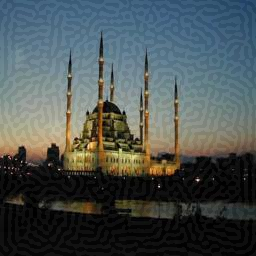}
        \caption*{Gabor~\cite{kang2019testing}}
    \end{subfigure} & 
    \begin{subfigure}[b]{0.16\textwidth}
        \includegraphics[width=\textwidth]{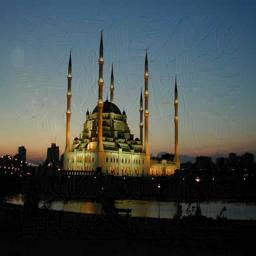}
        \caption*{PerC-AL~\cite{zhao2020towards}}
    \end{subfigure} \\
    \begin{subfigure}[t]{0.16\textwidth}
        \includegraphics[width=\textwidth]{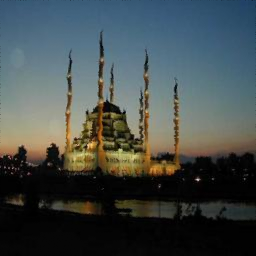}
        \caption*{StAdv~\cite{xiao2018spatially}}
    \end{subfigure} & 
    \begin{subfigure}[t]{0.16\textwidth}
        \includegraphics[width=\textwidth]{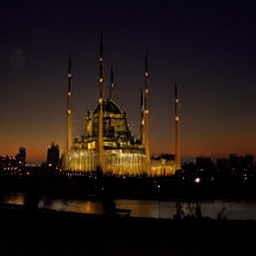}
        \caption*{\textbf{I2A:} \textit{``make it at night"}}
    \end{subfigure} & 
    \begin{subfigure}[t]{0.16\textwidth}
        \includegraphics[width=\textwidth]{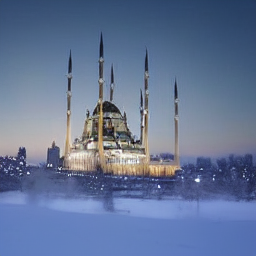}
        \caption*{\textbf{I2A:} \textit{``make it in snow"}}
    \end{subfigure} & 
    \begin{subfigure}[t]{0.16\textwidth}
        \includegraphics[width=\textwidth]{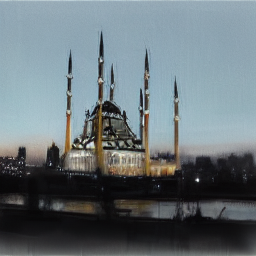}
        \caption*{\textbf{I2A:} \textit{``make it a sketch painting"}}
    \end{subfigure} & 
    \begin{subfigure}[t]{0.16\textwidth}
        \includegraphics[width=\textwidth]{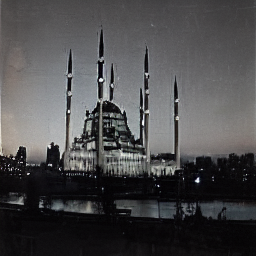}
        \caption*{\textbf{I2A:} \textit{``make it a vintage photo"}}
    \end{subfigure} & 
    \begin{subfigure}[t]{0.16\textwidth}
        \includegraphics[width=\textwidth]{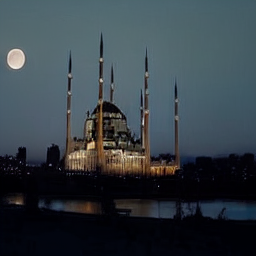}
        \caption*{\textbf{I2A:} \textit{``Add a full moon."} (GPT-4)}
    \end{subfigure} \\
\end{tabular}} 
\caption{\textbf{Visualization of adversarial images on Places365.} I2A generates natural and diverse perturbations based on the text instructions.}
\label{fig:vis_places}
\end{figure*}

\begin{figure*}[t]
\centering
\small
\setlength{\tabcolsep}{1mm}
\scalebox{0.95}{
\begin{tabular}[b]{ cccccc}
    \begin{subfigure}{0.16\textwidth}
        \includegraphics[width=\textwidth]{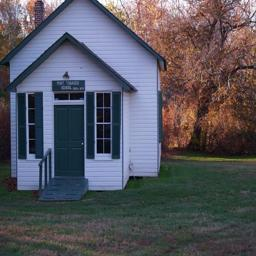}
        \caption*{Clean Image}
    \end{subfigure} & 
    \begin{subfigure}{0.16\textwidth}
        \includegraphics[width=\textwidth]{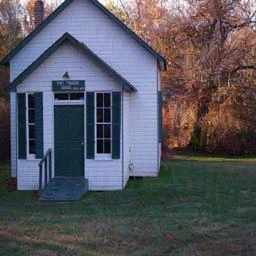}
        \caption*{PGD~\cite{madry2018towards}}
    \end{subfigure} & 
    \begin{subfigure}{0.16\textwidth}
        \includegraphics[width=\textwidth]{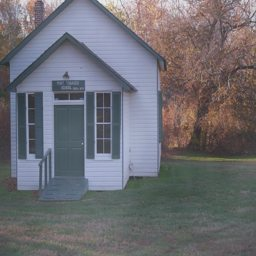}
        \caption*{Fog~\cite{kang2019testing}}
    \end{subfigure} & 
    \begin{subfigure}{0.16\textwidth}
        \includegraphics[width=\textwidth]{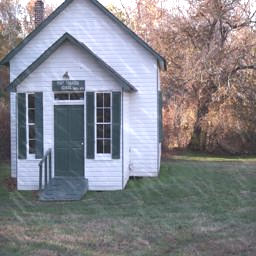}
        \caption*{Snow~\cite{kang2019testing}}
    \end{subfigure} & 
    \begin{subfigure}{0.16\textwidth}
        \includegraphics[width=\textwidth]{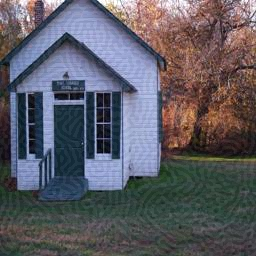}
        \caption*{Gabor~\cite{kang2019testing}}
    \end{subfigure} & 
    \begin{subfigure}{0.16\textwidth}
        \includegraphics[width=\textwidth]{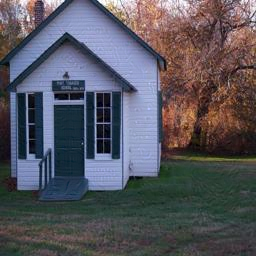}
        \caption*{PerC-AL~\cite{zhang2018perceptual}}
    \end{subfigure} \\
    \begin{subfigure}[t]{0.16\textwidth}
        \includegraphics[width=\textwidth]{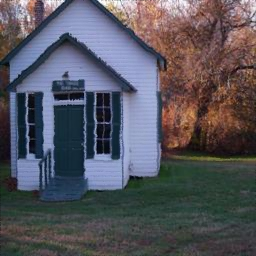}
        \caption*{StAdv~\cite{xiao2018spatially}}
    \end{subfigure} & 
    \begin{subfigure}[t]{0.16\textwidth}
        \includegraphics[width=\textwidth]{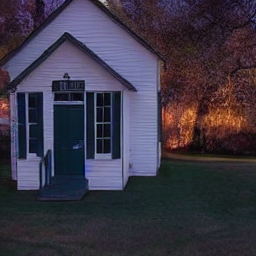}
        \caption*{\textbf{I2A:} \textit{``make it at night"}}
    \end{subfigure} & 
    \begin{subfigure}[t]{0.16\textwidth}
        \includegraphics[width=\textwidth]{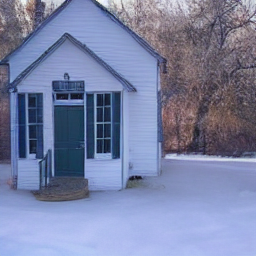}
        \caption*{\textbf{I2A:} \textit{``make it in snow"}}
    \end{subfigure} & 
    \begin{subfigure}[t]{0.16\textwidth}
        \includegraphics[width=\textwidth]{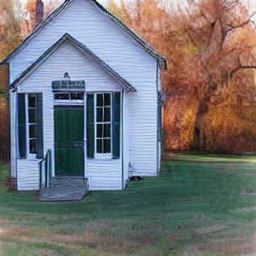}
        \caption*{\textbf{I2A:} \textit{``make it a sketch painting"}}
    \end{subfigure} & 
    \begin{subfigure}[t]{0.16\textwidth}
        \includegraphics[width=\textwidth]{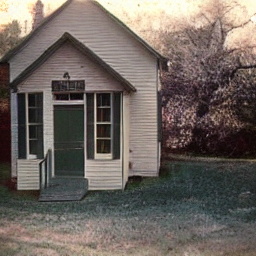}
        \caption*{\textbf{I2A:} \textit{``make it a vintage photo"}}
    \end{subfigure} & 
    \begin{subfigure}[t]{0.16\textwidth}
        \includegraphics[width=\textwidth]{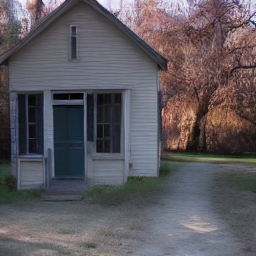}
        \caption*{\textbf{I2A:} \textit{``Add a dirt path leading to it."} (GPT-4)}
     \end{subfigure}  \\ \hline 
    \vspace{-5pt}\\
    \begin{subfigure}[b]{0.16\textwidth}
        \includegraphics[width=\textwidth]{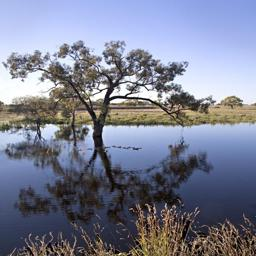}
        \caption*{Clean Image}
    \end{subfigure} & 
    \begin{subfigure}[b]{0.16\textwidth}
        \includegraphics[width=\textwidth]{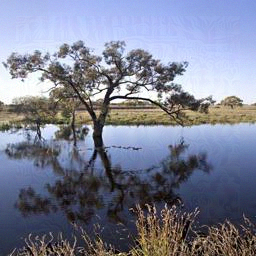}
        \caption*{PGD~\cite{madry2018towards}}
    \end{subfigure} & 
    \begin{subfigure}[b]{0.16\textwidth}
        \includegraphics[width=\textwidth]{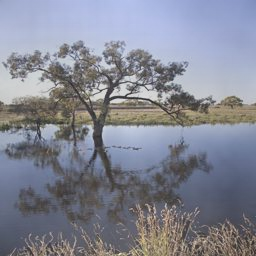}
        \caption*{Fog~\cite{kang2019testing}}
    \end{subfigure} & 
    \begin{subfigure}[b]{0.16\textwidth}
        \includegraphics[width=\textwidth]{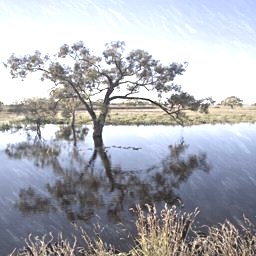}
        \caption*{Snow~\cite{kang2019testing}}
    \end{subfigure} & 
    \begin{subfigure}[b]{0.16\textwidth}
        \includegraphics[width=\textwidth]{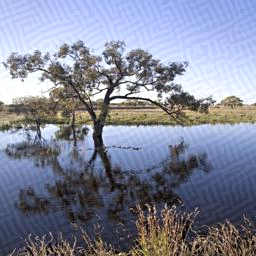}
        \caption*{Gabor~\cite{kang2019testing}}
    \end{subfigure} & 
    \begin{subfigure}[b]{0.16\textwidth}
        \includegraphics[width=\textwidth]{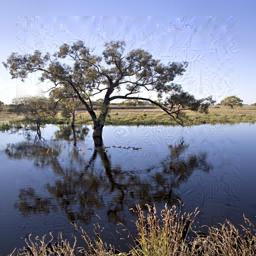}
        \caption*{PerC-AL~\cite{zhao2020towards}}
    \end{subfigure} \\
    \begin{subfigure}[t]{0.16\textwidth}
        \includegraphics[width=\textwidth]{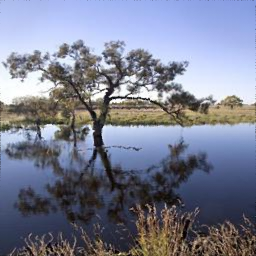}
        \caption*{StAdv~\cite{xiao2018spatially}}
    \end{subfigure} & 
    \begin{subfigure}[t]{0.16\textwidth}
        \includegraphics[width=\textwidth]{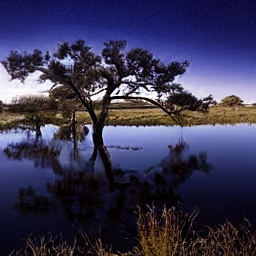}
        \caption*{\textbf{I2A:} \textit{``make it at night"}}
    \end{subfigure} & 
    \begin{subfigure}[t]{0.16\textwidth}
        \includegraphics[width=\textwidth]{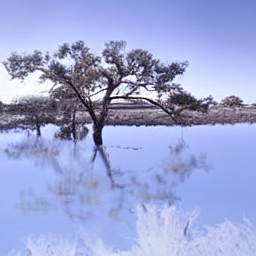}
        \caption*{\textbf{I2A:} \textit{``make it in snow"}}
    \end{subfigure} & 
    \begin{subfigure}[t]{0.16\textwidth}
        \includegraphics[width=\textwidth]{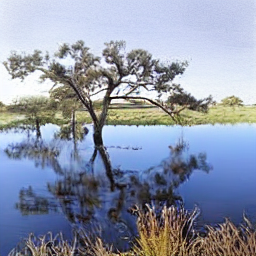}
        \caption*{\textbf{I2A:} \textit{``make it a sketch painting"}}
    \end{subfigure} & 
    \begin{subfigure}[t]{0.16\textwidth}
        \includegraphics[width=\textwidth]{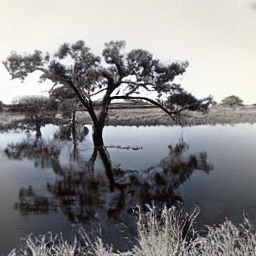}
        \caption*{\textbf{I2A:} \textit{``make it a vintage photo"}}
    \end{subfigure} & 
    \begin{subfigure}[t]{0.16\textwidth}
        \includegraphics[width=\textwidth]{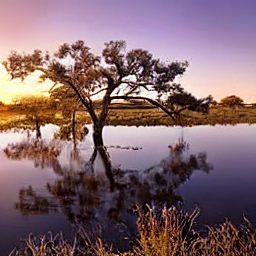}
        \caption*{\textbf{I2A:} \textit{``Make it sunset time"} (GPT-4)}
    \end{subfigure} \\ \hline 
    \vspace{-5pt}\\
    \begin{subfigure}[b]{0.16\textwidth}
        \includegraphics[width=\textwidth]{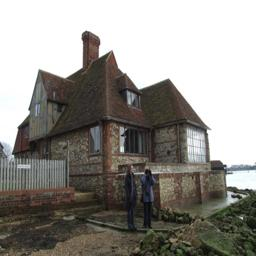}
        \caption*{Clean Image}
    \end{subfigure} & 
    \begin{subfigure}[b]{0.16\textwidth}
        \includegraphics[width=\textwidth]{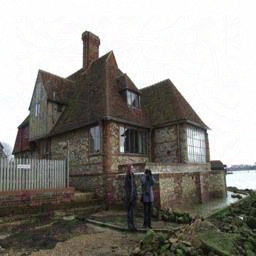}
        \caption*{PGD~\cite{madry2018towards}}
    \end{subfigure} & 
    \begin{subfigure}[b]{0.16\textwidth}
        \includegraphics[width=\textwidth]{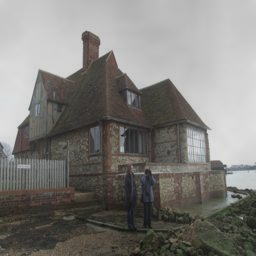}
        \caption*{Fog~\cite{kang2019testing}}
    \end{subfigure} & 
    \begin{subfigure}[b]{0.16\textwidth}
        \includegraphics[width=\textwidth]{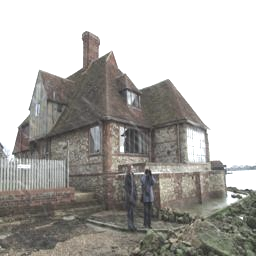}
        \caption*{Snow~\cite{kang2019testing}}
    \end{subfigure} & 
    \begin{subfigure}[b]{0.16\textwidth}
        \includegraphics[width=\textwidth]{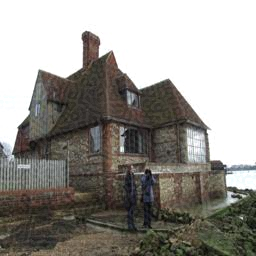}
        \caption*{Gabor~\cite{kang2019testing}}
    \end{subfigure} & 
    \begin{subfigure}[b]{0.16\textwidth}
        \includegraphics[width=\textwidth]{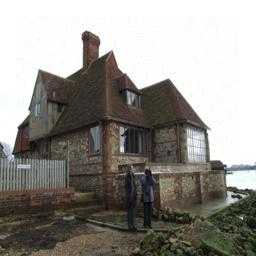}
        \caption*{PerC-AL~\cite{zhao2020towards}}
    \end{subfigure} \\
    \begin{subfigure}[t]{0.16\textwidth}
        \includegraphics[width=\textwidth]{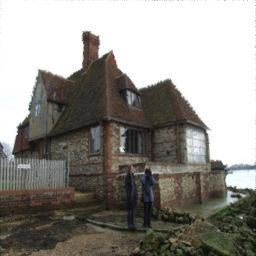}
        \caption*{StAdv~\cite{xiao2018spatially}}
    \end{subfigure} & 
    \begin{subfigure}[t]{0.16\textwidth}
        \includegraphics[width=\textwidth]{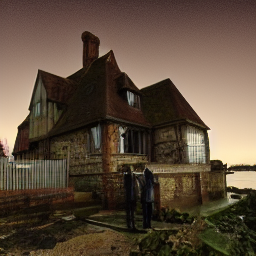}
        \caption*{\textbf{I2A:} \textit{``make it at night"}}
    \end{subfigure} & 
    \begin{subfigure}[t]{0.16\textwidth}
        \includegraphics[width=\textwidth]{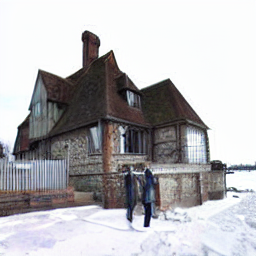}
        \caption*{\textbf{I2A:} \textit{``make it in snow"}}
    \end{subfigure} & 
    \begin{subfigure}[t]{0.16\textwidth}
        \includegraphics[width=\textwidth]{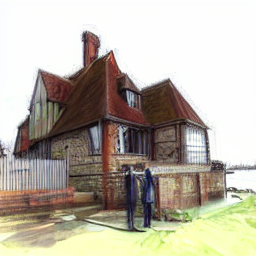}
        \caption*{\textbf{I2A:} \textit{``make it a sketch painting"}}
    \end{subfigure} & 
    \begin{subfigure}[t]{0.16\textwidth}
        \includegraphics[width=\textwidth]{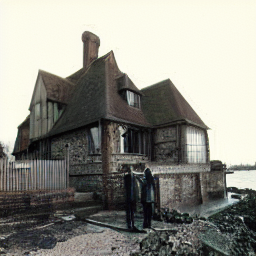}
        \caption*{\textbf{I2A:} \textit{``make it a vintage photo"}}
    \end{subfigure} & 
    \begin{subfigure}[t]{0.16\textwidth}
        \includegraphics[width=\textwidth]{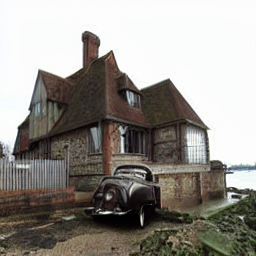}
        \caption*{\textbf{I2A:} \textit{``Add a vintage car."} (GPT-4)}
    \end{subfigure} \\
\end{tabular}} 
\caption{\textbf{Visualization of adversarial images on Places365.} I2A generates natural and diverse perturbations based on the text instructions.}
\label{fig:vis_places2}
\end{figure*}

We provide more visualization results on ImageNet~\cite{deng2009imagenet} in~\cref{fig:vis_supp-2,fig:vis_supp}. We also provide visualization results on Places365~\cite{zhou2017places} in~\cref{fig:vis_places,fig:vis_places2}. I2A generates diverse and natural adversarial images based on the language instructions, and generalizes well to different datasets.

\section{Prompt for GPT-4}
In this work, we propose to automatically generate image-specific editing instructions by first using BLIP-2~\cite{li2023blip} to generate image captions and feed them into GPT-4~\cite{openai2023gpt}. We utilize few-shot in-context learning by presenting 50 manually written examples in the prompt without fine-tuning the GPT-4 model. The prompt for GPT-4 to generate the edit instructions is presented in~\Cref{tab:gpt_prompt}.
\begin{table*}[h]\centering
\begin{minipage}{1.0\textwidth}\vspace{0mm}    \centering
\begin{tcolorbox} 
You are now tasked with generating image editing instructions for an advanced image editing algorithm. When given an image caption, your role is to produce a corresponding image editing instruction without altering the inherent nature or category of objects within the image. \\

\textbf{Guidelines:}
\begin{enumerate}
    \item  Do not alter the primary category of objects. For instance, if the caption mentions a ``beer glass", avoid instructions that would change it to a ``wine glass".
    \item Preserve the natural and typical attributes of objects. Hence, if the caption mentions a ``green mamba", don't instruct to change its color, given that green mambas are characteristically green.
    \item  Ensure that the resulting scene remains plausible. For example, if the caption says, ``a young orangutan standing on a rock", refrain from suggesting changes like ``make the rock green", which would create an unnatural scenario.  
    \item Refrain from introducing drastic global alterations like changing indoor scenes to outdoor.
    \item Be aware of potential errors in the captions and adhere to the object ``category".
    \item Prioritize Simplicity. Keep the edits straightforward and uncomplicated. \\
\end{enumerate}

\textbf{Here are some examples: }

\{``input": ``a close up of a gun on a soldier's shoulder", ``edit": ``as if it were a drawing", ``output": ``a drawing of a gun on a soldier's shoulder", ``category": ``assault rifle"\} \\
\{``input": ``a small chinchilla is being fed a toothbrush", ``edit": ``make the chinchilla white", ``output": ``a white chinchilla is being fed by a syringe", ``category": ``syringe"\} \\
\{``input": ``a close up of a spider with orange legs", ``edit": ``as a painting", ``output": ``a painting of close up of a cockroach with orange legs", ``category": ``cockroach"\} \\
\{``input": ``an old theater curtain with a light shining through it", ``edit": ``turn off the light", ``output": ``an old theater curtain without light shining through it", ``category": ``theater curtain"\}\\
\{``input": ``a laptop computer sitting on a desk with a keyboard", ``edit": ``make it black", ``output": ``a black laptop computer sitting on a desk with a keyboard", ``category": ``notebook"\}\\
\{``input": ``a dog standing on a dirt road next to a pole", ``edit": ``make it at night", ``output": ``a dog standing on a dirt road next to a pole during the night", ``category": ``malinois"\}\\
\{``input": ``a black and white photo of a spiral garden", ``edit": ``make it more colourful", ``output": ``a colourful photo of a spiral garden", ``category": ``maze"\}\\
\{``input": ``a close up of a curtain with a pattern on it", ``edit": ``make it blue", ``output": ``a close up of a blue curtain with a pattern on it", ``category": ``shower curtain"\}\\
\{``input": ``a large dinosaur statue with a big mouth", ``edit": ``add some snow", ``output": ``a large dinosaur statue with a big mouth in snow", ``category": ``triceratops"\}\\
\{``input": ``a bug on the hood of a car", ``edit": ``on Mars", ``output": ``a bug on the hood of a car on Mars", ``category": ``walking stick"\}\\
\{``input": ``a woman wearing a black jacket next to a vending machine", ``edit": ``change the jacket to a cape", ``output": ``a woman wearing a black cape next to a vending machine", ``category": ``vending machine"\}\\
\{``input": ``a brown and tan pitcher with a handle on a table", ``edit": ``Make it look like an old photograph", ``output": ``an old photograph of a brown and tan pitcher with a handle on a table", ``category": ``pitcher"\}\\
\{``input": ``a metal gate with a shadow on it", ``edit": ``as if it was a painting", ``output": ``a painting of a metal gate with a shadow on it", ``category": ``turnstile"\}\\
{......} \\
\{``input": ``a green jeep parked in front of a white building", ``edit": ``make the jeep red", ``output": ``a red jeep parked in front of a white building", ``category": ``jeep"\} \\

\textbf{Please write edits for the following samples:}

\{``input": ``a river flowing through a valley with snow capped mountains", ``category": ``valley", ``edit": ``", ``output": ``"\} \\
...... \\
\end{tcolorbox}
\vspace{-2mm}
\caption{\textbf{Prompt for GPT-4 to generate edit instructions.} We provide 50 manually written examples and ask GPT-4 to generate the edits for the other images given the image captions and object categories as input.}
    \label{tab:gpt_prompt}
\end{minipage}
\end{table*}

\end{document}